\documentclass[letterpaper]{article} 
\usepackage[]{aaai2026}  
\usepackage{times}  
\usepackage{helvet}  
\usepackage{courier}  
\usepackage[hyphens]{url}  
\usepackage{graphicx} 
\usepackage{natbib}  
\usepackage{caption} 
\usepackage[mathscr]{eucal}
\usepackage[cmex10]{amsmath}
\usepackage{epsfig,epsf,psfrag}
\usepackage{amssymb,amsmath,amsthm,amsfonts,latexsym}
\usepackage{amsmath,graphicx,bm,xcolor,url,overpic}
\usepackage{fixltx2e}
\usepackage{array}
\usepackage{verbatim}
\usepackage{bm}
\usepackage{algorithmic}
\usepackage{algorithm}
\usepackage{verbatim}
\usepackage{textcomp}
\usepackage{epstopdf}
\usepackage{subfigure}
\usepackage{acronym}
\usepackage{diagbox}
\usepackage{tikz}
\usetikzlibrary{graphs, positioning, quotes, shapes.geometric}

\newcommand{\openone}{\leavevmode\hbox{\small1\normalsize\kern-.33em1}} 

\catcode`~=11 \def\UrlSpecials{\do\~{\kern -.15em\lower .7ex\hbox{~}\kern .04em}} \catcode`~=13 

\allowdisplaybreaks[4]

\def\tv{\mathop{\mathrm{TV}}}

\newcommand{\calA}{\mathcal{A}}

\newcommand{\calP}{\mathcal{P}}

\newcommand{\calT}{\mathcal{T}}

\newcommand{\calX}{\mathcal{X}}

\newcommand{\calZ}{\mathcal{Z}}

\newcommand{\bbE}{\mathbb{E}}

\newcommand{\bbI}{\mathbb{I}}

\newcommand{\bbP}{\mathbb{P}}

\newcommand{\bbR}{\mathbb{R}}

\DeclareMathAlphabet{\mathbsf}{OT1}{cmss}{bx}{n}
\DeclareMathAlphabet{\mathssf}{OT1}{cmss}{m}{sl}

\DeclareSymbolFont{bsfletters}{OT1}{cmss}{bx}{n}  
\DeclareSymbolFont{ssfletters}{OT1}{cmss}{m}{n}
\DeclareMathSymbol{\bsfGamma}{0}{bsfletters}{'000}
\DeclareMathSymbol{\ssfGamma}{0}{ssfletters}{'000}
\DeclareMathSymbol{\bsfDelta}{0}{bsfletters}{'001}
\DeclareMathSymbol{\ssfDelta}{0}{ssfletters}{'001}
\DeclareMathSymbol{\bsfTheta}{0}{bsfletters}{'002}
\DeclareMathSymbol{\ssfTheta}{0}{ssfletters}{'002}
\DeclareMathSymbol{\bsfLambda}{0}{bsfletters}{'003}
\DeclareMathSymbol{\ssfLambda}{0}{ssfletters}{'003}
\DeclareMathSymbol{\bsfXi}{0}{bsfletters}{'004}
\DeclareMathSymbol{\ssfXi}{0}{ssfletters}{'004}
\DeclareMathSymbol{\bsfPi}{0}{bsfletters}{'005}
\DeclareMathSymbol{\ssfPi}{0}{ssfletters}{'005}
\DeclareMathSymbol{\bsfSigma}{0}{bsfletters}{'006}
\DeclareMathSymbol{\ssfSigma}{0}{ssfletters}{'006}
\DeclareMathSymbol{\bsfUpsilon}{0}{bsfletters}{'007}
\DeclareMathSymbol{\ssfUpsilon}{0}{ssfletters}{'007}
\DeclareMathSymbol{\bsfPhi}{0}{bsfletters}{'010}
\DeclareMathSymbol{\ssfPhi}{0}{ssfletters}{'010}
\DeclareMathSymbol{\bsfPsi}{0}{bsfletters}{'011}
\DeclareMathSymbol{\ssfPsi}{0}{ssfletters}{'011}
\DeclareMathSymbol{\bsfOmega}{0}{bsfletters}{'012}
\DeclareMathSymbol{\ssfOmega}{0}{ssfletters}{'012}

\newcommand{\tilc}{\tilde{c}}

\newcommand{\tilf}{\tilde{f}}

\newcommand{\tilG}{\tilde{G}}

\newcommand{\hatP}{\hat{P}}

\newcommand{\tilx}{\tilde{x}}
\newcommand{\tilX}{\tilde{X}}

\newcommand{\barr}{\bar{r}}

\newcommand{\barP}{\bar{P}}

\newcommand{\tilomega}{\tilde{\omega}}

\def\##1\#{\begin{align}#1\end{align}}
\def\$#1\${\begin{align*}#1\end{align*}}

\usepackage{thmtools, thm-restate}
\newtheorem{theorem}{Theorem} 
\newtheorem{lemma}{Lemma}

\newtheorem{proposition}[lemma]{Proposition}

\newtheorem{assumption}{Assumption}
\usepackage{amsmath,amssymb,amsfonts}
\usepackage{algorithmic}
\usepackage{graphicx}
\usepackage{textcomp}
\def\BibTeX{{\rm B\kern-.05em{\sc i\kern-.025em b}\kern-.08em
    T\kern-.1667em\lower.7ex\hbox{E}\kern-.125emX}}
\newcommand{\pt}{\texttt{pt}}
\newcommand{\nor}{\mathrm{Norm}}
\newcommand{\van}{\mathrm{van}}

\newcommand{\ourmethod}{\textsc{Cool-SD}}
\newcommand{\unirelax}{\textsc{UniformRSD}}
\newcommand{\unif}{\text{Uniform}}

\newcommand{\vanilla}{\colorbox{pink}{{\textbf{Vanilla SD}}}}
\newcommand{\ourrelaxed}{\colorbox{lightgreen}{{\textbf{\ourmethod}\vphantom{Vg}}}}
\usepackage{mathtools}
\mathtoolsset{showonlyrefs}

\usepackage{acronym}
\acrodef{ar}[AR]{Auto-Regressive}
\acrodef{sd}[SD]{Speculative Decoding}
\acrodef{llm}[LLM]{Large Language Models}

\usepackage{algorithm}
\usepackage{algorithmic}
\usepackage{booktabs} 
\usepackage{multirow}

\definecolor{lightgreen}{RGB}{230, 254, 238}  
\definecolor{pink}{RGB}{252, 238, 240}        
\definecolor{darkred}{RGB}{179, 0, 0}
\definecolor{darkgreen}{RGB}{0, 89, 0}
\usepackage{calc} 

\newcommand{\addedline}[1]{\colorbox{lightgreen}{\parbox{\dimexpr\linewidth-2\fboxsep}{{#1}}}}
\newcommand{\deletedline}[1]{\colorbox{pink}{\parbox{\dimexpr\linewidth-2\fboxsep}{{#1}}}}
\newcommand{\addedlineb}[1]{\colorbox{lightgreen}{\parbox{\dimexpr\linewidth-4.8\fboxsep}{{#1}}}}
\newcommand{\deletedlineb}[1]{\colorbox{pink}{\parbox{\dimexpr\linewidth-4.8\fboxsep}{{#1}}}}

\usepackage{newfloat}
\usepackage{listings}
\DeclareCaptionStyle{ruled}{labelfont=normalfont,labelsep=colon,strut=off} 
\floatstyle{ruled}
\newfloat{listing}{tb}{lst}{}
\floatname{listing}{Listing}
\title{Annealed Relaxation of Speculative Decoding \\for Faster Autoregressive Image Generation}
\author {
    Xingyao Li\textsuperscript{\rm 1}, 
    Fengzhuo Zhang\textsuperscript{\rm 1}, 
    Cunxiao Du\textsuperscript{\rm 2}\thanks{Corresponding author. },
    Hui Ji\textsuperscript{\rm 1}
}
\affiliations {
    \textsuperscript{\rm 1}National University of Singapore\\ \textsuperscript{\rm 2}Sea AI Lab\\
    xingyao@u.nus.edu, \, fzzhang@u.nus.edu, \, cnsdunm@gmail.com, \, matjh@nus.edu.sg
}

\usepackage{bibentry}

\begin{document}
\maketitle

\begin{abstract}

Despite significant progress in auto-regressive  image generation, inference remains slow due to the sequential nature of AR models and the ambiguity of image tokens, even when using speculative decoding. Recent works attempt to address this with relaxed speculative decoding but lack theoretical grounding. In this paper, we establish the theoretical basis of relaxed SD and propose \ourmethod{}, an annealed relaxation of speculative decoding built on two key insights. The first analyzes the total variation (TV) distance between the target model and relaxed speculative decoding and yields an optimal resampling distribution that minimizes an upper bound of the distance. The second uses perturbation analysis to reveal an annealing behaviour in relaxed speculative decoding, motivating our annealed design. Together, these insights enable \ourmethod{} to generate images faster with comparable quality, or achieve better quality at similar latency. Experiments validate the effectiveness of \ourmethod{}, showing consistent improvements over prior methods in speed-quality trade-offs.

\end{abstract}

\begin{links}
    \link{Code}{https://github.com/xingyaoL/COOL-SD}
\end{links}
\section{Introduction}

\ac{ar} models have recently emerged as a powerful paradigm for image generation, often achieving performance comparable to or even surpassing that of diffusion-based methods~\citep{sun2024autoregressive,chen2025janus,rombach2022high,esser2024scaling}. These models generate images in a sequential manner  by predicting one token at a time, conditioned on all previously generated tokens. While this \ac{ar} factorization is effective for modeling complex dependencies, it leads to substantial inference latency. Each token requires a separate forward pass, making the decoding process very costly for high-resolution images with  thousands of tokens. Consequently, \ac{ar} models are inefficient for real-time or interactive applications. Accelerating the decoding process is therefore a key challenge for making AR models more practical and widely applicable.

\begin{figure}[t]
    \centering
    \includegraphics[width=\linewidth]{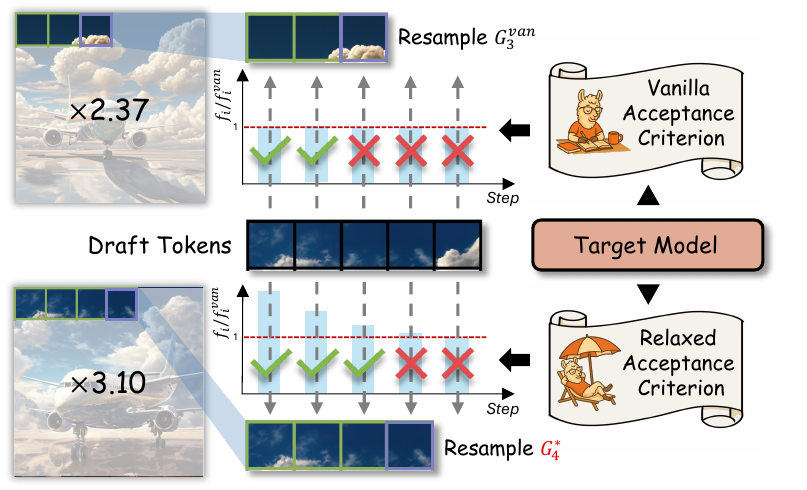}
    \caption{{Illustration of vanilla SD and \ourmethod{}}. By increasing the acceptance criterion $f_{i}$ according to an annealing schedule, along with a principled resampling distribution $G_{i}^{*}$, we can further increase the inference speed. }
    \label{fig:illustration}
\end{figure} 

\subsubsection{Speculative Decoding and Its Variants}
\ac{sd} ~\citep{leviathan2023fast,chen2023accelerating}
has emerged as an effective lossless approach for reducing the latency of auto-regressive generation, producing outputs from the exact target distribution while requiring fewer forward passes. \ac{sd} leverages two models: a high-quality but slow target model and a smaller, faster draft model. The decoding process proceeds in two stages: drafting, where the draft model auto-regressively generates a sequence of tokens, and verification, where the target model evaluates the entire draft sequence in a single forward pass and accepts a prefix of tokens that match its predictions. If any token is rejected, the target model resamples to correct for distributional bias. The accepted tokens are appended to the prefix, and the process repeats. By reducing the number of calls to the target model, \ac{sd} accelerates the process without altering the output distribution.

\begin{figure*}[t]
    \centering
    \includegraphics[width=0.77\textwidth]{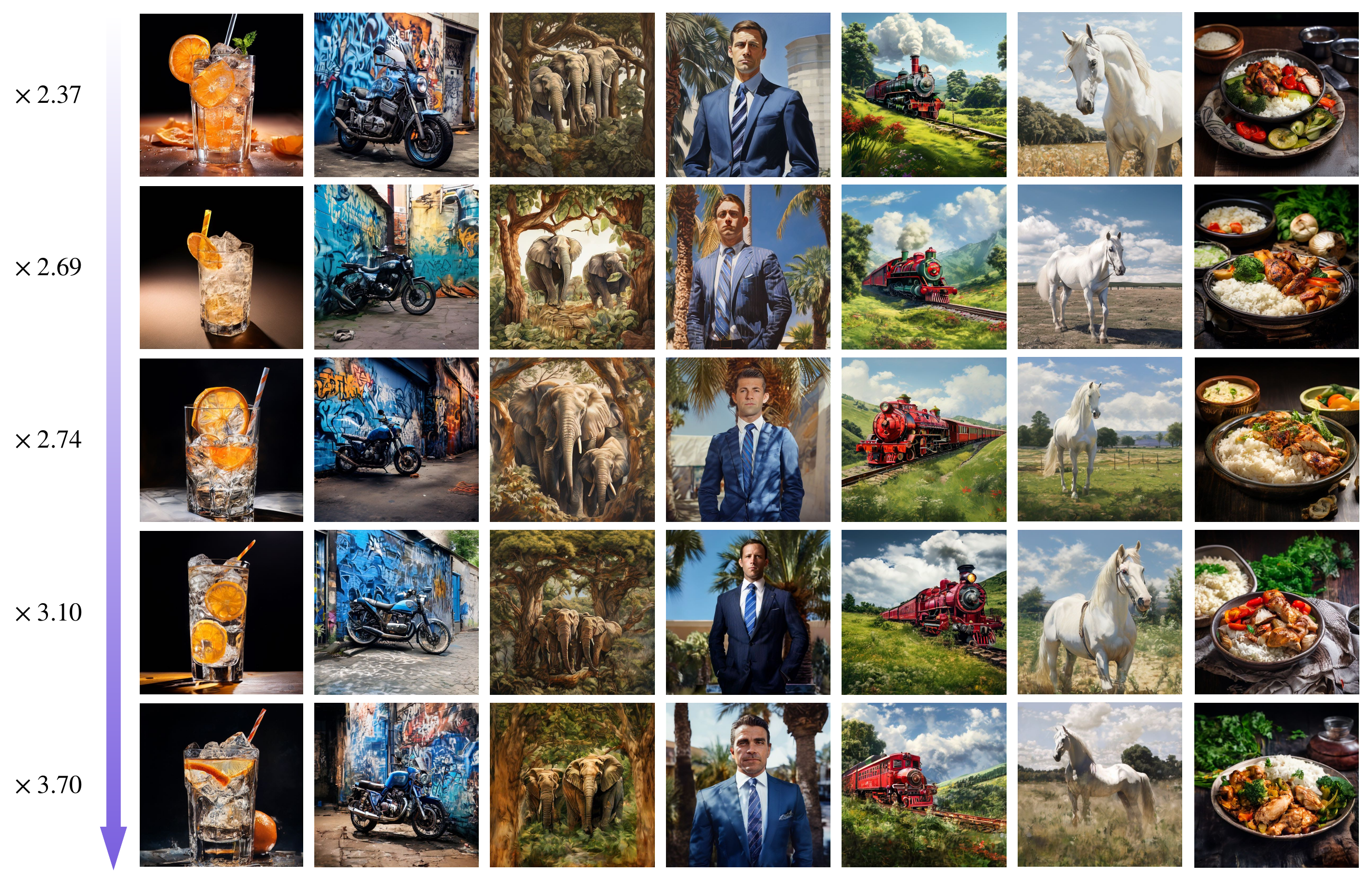}
    \caption{{Qualitative results of \ourmethod{}}. We demonstrate the trade-off between generation efficiency and image quality by comparing outputs from \ourmethod{} on Lumina-mGPT under different parameter settings. The speed-up factor is annotated to the left of each row. The first row shows images generated by Eagle-1 without any relaxation, serving as the baseline.}\label{fig:visual}
\end{figure*}

Despite its success in accelerating the \ac{llm} inference~\citep{cai2024medusa,li2024eagle}, \ac{sd}'s acceleration effect in image generation is  limited. Prior studies attribute this to token selection ambiguity~\citep{jang2024lantern}. That is, at many decoding steps, multiple candidate tokens receive nearly identical probabilities, increasing the difficulty for the draft model to accurately predict the target model’s top-ranked tokens. This mismatch lowers the token acceptance rate during the verification stage, thereby reducing its overall efficiency. There have been several attempts to address this issue. LANTERN~\citep{jang2024lantern} introduces a relaxed verification scheme that permits limited divergence from the target distribution, while LANTERN++~\citep{park2025lantern++} improves efficiency via a static tree structure.  However, both rely on heuristic, empirically driven designs and  lack any mathematical account of their divergence from the target distribution. Moreover, their performance gains remain modest, with room for improvement in either  image quality or generation speed. This calls for a more principled approach to relaxed \ac{sd},  with  mathematical analysis on the approximation to the target distribution, to achieve  better   quality  at the same level of acceleration.

\subsubsection{Main Idea and Contributions}
Driven by the need for a principled and effective relaxation of speculative decoding, this paper aims at developing an acceleration scheme  that improves the speed-quality trade-off beyond vanilla \ac{sd}, supported by rigorous analysis of distributional fidelity.

\vspace{-0.2em}
In this paper, we propose an acceleration technique that builds on a generalization of vanilla \ac{sd}, replacing its rigid, exact-match acceptance test with a relaxed alternative. In our relaxed framework, each drafted token is accepted with a controllable probability $f_i$, rather than only when it exactly matches the target model's prediction. While this decouples the drafting and verification stages, thus enabling longer expected prefixes, it also introduces distributional drift. To address this, we derive token-wise resampling distributions $\{G_i^*\}_{i=1}^{L}$ by minimizing an almost-tight upper bound on the total variation (TV) distance between the output distribution and the target model. These optimal resampling distributions are specifically designed to closely approximate the target distribution under relaxed acceptance, and can be seamlessly integrated into any relaxed \ac{sd} variant.

Furthermore, we showed that  an \emph{annealing property} of the acceptance criterion: for a fixed expected number of accepted tokens, the bound is minimized when acceptance probabilities decrease monotonically across positions. Motivated by this, we adopt an exponentially decaying acceptance schedule in relaxing $f_i$, and combine it with the our derived resampling distributions $\{G_i^*\}$. This leads to a more efficient and theoretically grounded decoding strategy, which we refer to as \textbf{\ourmethod{}}.

 In short,  we present a theoretically grounded and empirically effective framework for accelerating AR  generation via a principled relaxation of \ac{sd}. Our main contributions are:
 
    \noindent
    $\bullet$ We revisit  \ac{sd}  and study its relaxed framework that replaces the rigid exact-match acceptance test with general acceptance rules.
    Within this new framework, we present several insights for algorithm design of relaxed \ac{sd}. 
    Specifically, we derive the expected number of accepted tokens and a nearly-tight upper bound on the TV distance.
    Minimizing this bound yields token-wise optimal resampling distributions that closely approximate the target model. We further show that a monotonically decaying acceptance schedule minimizes the bound.
    
    \noindent
    $\bullet$ Building on these insights, we propose a scheme which integrates the derived resampling distributions with an exponentially decaying acceptance schedule, namely \ourmethod{}. It consistently improves the speed-quality Pareto frontier over existing \ac{sd} variants and offers a principled foundation for future advances in efficient \ac{ar} image generation.
    
    \noindent
    $\bullet$ Extensive experiments on image generation tasks validate the effectiveness of our approach, showing consistent improvements over existing related techniques in both speed and quality trade-offs. Applying the derived optimal resampling distributions to the existing relaxed speculative decoding method, LANTERN, also yields significant gains in the efficiency and fidelity trade-off.

\section{Related Work}
\subsection{AR Image Generation}
 In addition to diffusion-based models~\citep{rombach2022high} and flow-matching methods~\citep{esser2024scaling}, auto-regressive generation of image tokens has recently emerged as a promising approach for unifying text and image generation~\citep{sun2024autoregressive,liu2024lumina}. LlamaGen~\citep{sun2024autoregressive} is the first to adopt and fintune the LLaMA base model for image generation by training a VQ-VAE image tokenizer~\citep{van2017neural, esser2021taming}. 
Building on this foundation, subsequent works have improved generation quality, modality coverage, and output diversity through better tokenizers, datasets, architectures, and training strategies~\citep{liu2024lumina,chen2025janus,wang2024emu3,team2024chameleon,xie2024show}. A notable advantage of these \ac{ar} models is their ability to generate \emph{multi-modal} tokens with a single model via a shared tokenizer~\citep{liu2024lumina}. A comprehensive survey can be found in~\citet{xiong2024autoregressive}.

\subsection{SD and Its Extensions} To mitigate the latency of \ac{ar} generation in \ac{llm}s, \citet{leviathan2023fast} and \citet{chen2023accelerating} propose Speculative Decoding (SD), which accelerates token generation by enabling a large model (the target model) to verify a sequence of draft tokens generated by a smaller model (the draft model). Follow-up works extend this linear-chain structure to tree-based variants for further acceleration~\citep{miao2023SpecInferAG,cai2024medusa,du2024glide,li2024eagle,huaccelerated}. SpecTr~\citep{sun2024spectr} formulates the problem from an optimal transport perspective and derives an algorithm that is optimal up to a multiplicative constant. \citet{sun2024block} propose verifying all draft tokens as a single block, further improving the acceleration ratio. \citet{yin2024theoretical} provide a theoretical analysis of \ac{sd}, while \citet{bachmann2025judge} and \citet{zhong2025speeding} explore classifier-based verification to accept more draft tokens and reduce verification time. Although these works primarily target text generation, \citet{jang2024lantern} adapt \ac{sd} to the image domain by relaxing the verification mechanism and propose LANTERN, significantly reducing generation latency. LANTERN++~\citep{park2025lantern++} further improves this approach by replacing the dynamic tree structure with a static one. However, the theoretical foundations of relaxed \ac{sd} for image generation remain largely unexplored in the current literature. More discussions on related methods are provided in Appendix~\ref{app:related_works}.

\section{Method}

\subsection{Speculative Decoding}

\ac{sd}~\citep{chen2023accelerating} is an acceleration technique for \ac{ar} generation that enables the simultaneous generation of multiple tokens. 
We present the vanilla \ac{sd} algorithm~\citep{chen2023accelerating,leviathan2023fast} in Algorithm~\ref{alg:approx_SD}. 
It involves two models: a target
model $P$, and a faster draft model $Q$.
In the context of \ac{ar} image generation,
the target model $P$ is typically a large image generation model such as LlamaGen~\citep{sun2024autoregressive} or Lumina-mGPT~\citep{liu2024lumina}. The draft model $Q$ is a significantly smaller neural network (thus faster), and is trained to predict the behavior of the target model $P$. Given the prefix $\pt$ to decode, which includes the input prompt and the previously generated tokens, the target (resp. draft) model predicts the next token by sampling from $P(\cdot\,|\,\pt)$ (resp. $Q(\cdot\,|\,\pt)$). One round of \ac{sd} consists of two main stages: the \emph{drafting} stage and the \emph{verification} stage.

\vskip 2pt

\noindent
1. \textbf{Drafting Stage}: In this stage, SD uses the draft model $Q$ to generate subsequent $L$ candidate tokens 
$\tilx_{i} \sim Q(\cdot\,|\,\pt,\tilx_{1:i-1})$ for $i \in [L]$ in an \ac{ar} manner, shown in Line~\ref{ln:draft} of Algorithm~\ref{alg:approx_SD}.

\begin{algorithm}[t]
	\caption{\vanilla ($-$) / \ourrelaxed ($+$)}
        \label{alg:approx_SD}
        \textbf{Input:} target model $P$, draft model $Q$, prefix $\pt$, draft length $L$.\\
	\textbf{Procedure:}
	\begin{algorithmic}[1]\label{algo:t_asd}
        \STATE \textit{ // Drafting Stage:}\\
            \STATE Generate $L$ draft tokens $\tilx_{1:L}$ via $\tilx_{i}\sim Q(\cdot\,|\,\pt,\tilx_{1:i-1})$ for $i\in[L]$. \label{ln:draft}\\
        \STATE \textit{ // Verification Stage:}\\
            \STATE Set $\tau=0$, and calculate the values of $P(\tilx_i\,|\,\pt,\tilx_{1:i-1})$ for $i\in[L]$ in parallel.\label{ln:parallel}\\
        \FOR{$i=1,\ldots,L$}
            \STATE Sample $r_i\sim\unif([0,1])$\\
            \STATE \deletedline{\textbf{if} $r_{i}\leq f_{i}^{\van}(\tilx_{1:i},P,Q)$ in Eqn.~\eqref{ieq:f_tsd} \textbf{then} \label{ln:accept}\hfill $-$}
            \STATE \addedline{\textbf{if} $r_{i}\leq f_{i}(\tilx_{1:i},P,Q;\omega_{i})$ in Eqn.~\eqref{eq:relax} \& \eqref{eq:sdar} \textbf{then} \hfill $+$}
            \STATE \hspace{0.7em} Set $\tau=i$ and $x_i=\tilx_i$. 
            \STATE  \textbf{else}
            \STATE \hspace{0.7em} \deletedlineb{Sample $x_i\sim G_{i}^{\van}(\tilx_{1:i-1},P,Q)$ in Eqn.~\eqref{eq:g_tsd} \label{ln:resample}\hfill $-$}
            \STATE \hspace{0.7em} \addedlineb{Sample $x_i\sim G_{i}^{*}(\tilx_{1:i-1},P,Q)$ in Eqn.~\eqref{eq:G_design} \hfill $+$}
            \STATE \hspace{0.7em} \textbf{break.}
            \STATE  \textbf{end if} 
        \ENDFOR
        \STATE \textbf{if} $\tau=L$ \textbf{then} sample $x_{L+1}\sim P(\cdot\,|\,\pt,x_{1:L})$ \textbf{end if}~\label{ln:bonus}
        \STATE Return $x_{1:\tau+1}$.
	\end{algorithmic}
\end{algorithm}

\vskip 2pt

\noindent
2. \textbf{Verification Stage}: The target model $P$ first computes the corresponding token distributions $P(\cdot\,|\,\pt,\tilx_{1:i-1})$ in parallel for $i \in [L+1]$ (Line~\ref{ln:parallel}). After that, each drafted token $\tilx_i$ is independently accepted with probability (Line~\ref{ln:accept})
\begin{align}\label{ieq:f_tsd}
f_{i}^{\van}(\tilx_{1:i}, P, Q) := 
\min\left\{1,\frac{P(\tilx_{i}\,|\,\pt,\tilx_{1:i-1})}{Q(\tilx_{i}\,|\,\pt,\tilx_{1:i-1})}\right\}.
\end{align}
Let $\tau$ denote the number of accepted tokens out of the $L$ drafted candidates. 
If all drafted tokens are accepted, i.e. $\tau = L$, as the target model has already computed the sampling probability $P(\cdot\,|\,\pt,x_{1:L})$ for $x_{L+1}$ during the drafting stage, we can obtain a total of $L+1$ tokens in this \ac{sd} round~(Line~\ref{ln:bonus}). 
If a token is rejected at some position $\tau + 1$, the distribution will be corrected by sampling the $\tau + 1$-th token from the resampling distribution $G_{\tau+1}^{\van}(\tilx_{1:\tau}, P, Q)$ (Line~\ref{ln:resample}). $G_{\tau+1}^{\van}(\tilx_{1:\tau}, P, Q)$ is carefully designed
to ensure that the generated tokens $\tilx_{1:\tau+1}$ follow the target model's distribution, defined as
\begin{align}
&G_{\tau+1}^{\van}\big(\cdot \,|\, (\pt,\tilx_{1:\tau}),P,Q\big)\nonumber \\*
&= \nor\Big(\big[P(\cdot\,|\,\pt,\tilx_{1:\tau}) - Q(\cdot\,|\,\pt,\tilx_{1:\tau})\big]_{+}\Big).\label{eq:g_tsd}
\end{align}

\noindent
Here $\nor(\cdot)$ normalizes the input to a probability distribution and $[\,\cdot\,]_+ $ is defined as $ \max\{0,\cdot\}.$
The choice of $G_{\tau+1}^{\van}$ effectively rectifies the bias introduced by the draft model, as presented in Theorem~\ref{thm:t_spec}.
\begin{theorem}[Speculative Decoding Recovers Unbiased Target Distribution~\citep{chen2023accelerating}]\label{thm:t_spec}
    The output tokens $x_{1:\tau+1}$ obtained with vanilla \ac{sd} follows the target model distribution $x_{1:\tau+1}\sim P(\cdot\,|\,\pt)$. 
\end{theorem}

While vanilla \ac{sd} is originally proposed for accelerating \ac{llm}s, \citet{jang2024lantern} discovered its inefficiency in \ac{ar} image generation and enhance inference efficiency by increasing the acceptance probability. 
LANTERN~\citep{jang2024lantern} leverages the latent proximity of image tokens by aggregating the probabilities of a candidate token’s $k$-nearest neighbors into the candidate’s own probability mass. Following this, LANTERN++~\citep{park2025lantern++} replaces dynamic tree-based \ac{sd} with a static tree structure, which demonstrates improved performance on image generation tasks, and refines the design of relaxation parameters. Specifically, LANTERN++~\citep{park2025lantern++} modify the candidate token's probability distribution to 
\begin{align}\label{ieq:p_lantern}
&P^{k,\lambda}(x\,|\,\pt,\tilx_{1:i-1})\\*
&\quad := \left\{
\begin{aligned}
    &\textstyle\sum_{x\in A^{k,\lambda}(\tilx_{i})}P(x\,|\,\pt,\tilx_{1:i-1}), \text{ if } x = \tilx_{i}, \\
    &\quad 0, \qquad\qquad\text{if } x\in A^{k,\lambda}(\tilx_{i}) \text{ and } x \ne \tilx_{i}, \\
    & P(x\,|\,\pt,\tilx_{1:i-1}), \qquad\qquad\quad\,\text{ otherwise. }
\end{aligned}\right.
\end{align}

\noindent
Here, $A^{k,\lambda}(\tilx_{i})$ denotes a selected subset in the $k$-nearest neighbors set of $\tilx_{i}$ that satisfy $\sum_{x\in A^{k,\lambda}(\tilx)\setminus\tilx}P(x\,|\,\pt,\tilx_{1:i-1})<\lambda P(\tilx\,|\,\pt,\tilx_{1:i-1})$. 

Accordingly, the acceptance criterion in Line~\ref{ln:accept} of Algorithm~\ref{alg:approx_SD} is redefined as 
\begin{align}\label{ieq:f_lantern}
f_{i}^{k,\lambda}(\tilx_{1:i}, P, Q) \!:= \!
\min\left\{1,\frac{\sum\limits_{x\in A_{k,\lambda}(\tilx_{i})}P(x\,|\,\pt,\tilx_{1:i-1})}{Q(\tilx_{i}\,|\,\pt,\tilx_{1:i-1})}\right\}, 
\end{align}

\vspace{-1.5em}
\noindent
and the corresponding resampling distribution in Line~\ref{ln:resample} is modified to be
\begin{align}
&G_{\tau+1}^{k,\lambda}\big(\cdot \,|\, (\pt,\tilx_{1:\tau}),P,Q\big)\nonumber \\
&:= \nor\Big(\big[P^{k,\lambda}(\cdot\,|\,\pt,\tilx_{1:\tau}) - Q(\cdot\,|\,\pt,\tilx_{1:\tau})\big]_{+}\Big). \label{eq:g_lantern}
\end{align}

\noindent
However, we argue that the choice of $\{f_{i}^{k,\lambda}\}_{i=1}^L$ and $\{G_{i}^{k,\lambda}\}_{i=1}^L$ lacks theoretical justification. Consequently, the fidelity of the images generated by LANTERN remains uncertain. In the following, we conduct theoretical analysis of relaxed \ac{sd} in its general form.

\subsection{Annealed Relaxation of \ac{sd}} 

A key limitation in improving the acceptance rate of vanilla \ac{sd} lies in the rigidity of its acceptance criterion, as defined in Eqn.~\eqref{ieq:f_tsd}. Relaxing this criterion can increase the likelihood of accepting drafted tokens, particularly beneficial for image generation due to the semantic ambiguity of image tokens~\citep{jang2024lantern}. However, the acceptance functions $f_{i}^{\van}$ and resampling distributions $G_{i}^{\van}$ proposed in \citet{chen2023accelerating} were shown to be optimal solutions to a maximal coupling problem~\citep{sun2024spectr}. Consequently, any relaxation of the acceptance criterion inevitably introduces bias that cannot be perfectly corrected by adjusting the resampling distributions.

\subsubsection{Relaxed Acceptance Criteria and Resampling Distribution}
Motivated by the aforementioned unavoidable bias, we analyze the trade-off between efficiency and fidelity loss in any relaxed \ac{sd} algorithm. To encompass a broad class of potential algorithms, we generalize the acceptance criteria (Line~\ref{ln:accept}) and the corresponding resampling distributions (Line~\ref{ln:resample}) in Algorithm~\ref{alg:approx_SD} as $\{f_i\}_{i=1}^L$ and $\{G_i\}_{i=1}^L$, respectively. We require only that each $f_i$ takes $\tilx_{1:i}$, $P$, and $G$, and each $G_i$ takes $\tilx_{1:i-1}$, $P$, and $G$ as inputs. We refer to this general framework as \emph{Relaxed} \ac{sd}.

To enable a principled trade-off between efficiency and fidelity, we begin by formally quantifying both concepts. We measure \emph{efficiency} by the expected number of accepted tokens, denoted as $\tau$. The \emph{fidelity loss} is evaluated using the total variation (TV) distance, a standard metric for quantifying the discrepancy between probability distributions. Specifically, the TV distance between two distributions $P_1$ and $P_2$ over a discrete support $\calX$ is defined as 
$$
    \tv(P_{1}, P_{2}) := \sum_{x\in\calX}|P_1(x)-P_2(x)|/2.
$$

The length of the generated token sequence $x_{1:\tau+1}$ is a random variable, influenced by the specific choices of the acceptance criteria $\{f_i\}_{i=1}^L$. 
For instance, a design with lower acceptance probabilities tends to yield shorter sequences (i.e., smaller $\tau$). 
Therefore, to enable a fair comparison of the fidelity loss across different designs, we \emph{virtually} extend each generated sequence to a fixed length of $L+1$ by sampling the remaining tokens $x_{\tau+2:L+1}$ from the target model, i.e., $x_i \sim P(\cdot \,|\, \pt, x_{1:i-1})$ for $i \in \{\tau+2, \ldots, L+1\}$. 
Since these additional tokens are generated from the target model $P$, this virtual extension does not introduce any extra fidelity loss. We denote the resulting full sequence distribution as $\hat{P}$, i.e., 
$
    x_{1:L+1} \sim \hatP(\cdot\,|\,\pt) := \hatP_{X_{1:L+1}}.  
$
For notational consistency, we denote the target distribution as $P(\cdot\,|,\pt) := P_{X_{1:L+1}}$.

We state our analysis in Theorem~\ref{thm:t_tvbd}.
For simplicity, we omit the notation for the prefixed tokens $\pt$ in $f_i$, $G_i$, $P$, and $Q$ throughout the remainder of the manuscript.

\begin{theorem}\label{thm:t_tvbd}
For the tokens generated by Relaxed \ac{sd} with acceptance criteria $\{f_i\}_{i=1}^L$ and resampling distributions $\{G_i\}_{i=1}^L$ in Algorithm~\ref{algo:t_asd}, the expectation of the number of accepted tokens is
    \begin{align}\bbE[\tau+1]=1+\sum_{i=1}^{L}\sum_{\tilx_{1:i}\in\calX^{i}}\!Q(\tilx_{1:i})\prod_{j=1}^{i}f_{j}(\tilx_{1:j},P,Q).\label{eq:token_exp}
    \end{align}
The total variation between $\hatP$ and $P$ is upper bounded as
    \begin{align}
        &\tv(\hatP_{X_{1:L+1}},P_{X_{1:L+1}})\label{ieq:tv_token}\\
        &\quad\leq \frac{1}{2}\sum_{i=0}^{L-1}\sum_{x_{1:i+1}\in\calX^{i+1}}Q(x_{1:i})\prod_{k=1}^{i}f_{k}(x_{1:k},P,Q)\cdot \nonumber\\
        &\qquad\bigg|Q(x_{i+1}\,|\, x_{1:i})\cdot f_{i+1}(x_{1:i+1},P,Q)-P(x_{i+1}\,|\, x_{1:i})\nonumber\\
        &\qquad +G_{i+1}(x_{i+1}\,|\,x_{1:i},P,Q)\nonumber\\
        &\qquad \cdot\!\!\!\sum_{\tilx_{i+1}\in\calX}\Big[1-f_{i+1}\big((x_{1:i},\tilx_{i+1}),P,Q\big)\Big]Q(\tilx_{i+1}\,|\,x_{1:i})\bigg|.
    \end{align}

    \vspace{-1.5em}
    \noindent
    Given $\{f_{i}\}_{i=1}^{L}$, one minimizer $\{G_{i}\}_{i=1}^{L}$ of this upper bound is
    $G_{i+1}^{*}(\cdot\mid \tilx_{1:i},P,Q)=$
    \vspace{-0.2em}
    \noindent
    \begin{align}
        \nor\Big(\!\big[\!P(\cdot| \tilx_{1:i})\!-\!Q(\cdot|  \tilx_{1:i}) f_{i+1}\big((\tilx_{1:i},\cdot),P,Q\big)\big]_{+}\Big).\label{eq:opt_G}
    \end{align}
\end{theorem}

The proof of Theorem~\ref{thm:t_tvbd} is provided in Appendix~\ref{app:proof_main}. The theorem first derives the expected accepted sequence length under $P$ and $Q$ in Eqn.~\eqref{eq:token_exp}, then establishes an upper bound on the total variation between the relaxed resulting distribution $\hatP$ and the target distribution $P$. This upper bound is almost-tight in the sense that when $f_{i}$ and $G_{i}$ are chosen as $f_{i}^{\van}$ and $G_{i}^{\van}$, the bound reduces to $0$. Finally, we derive the optimal resampling distribution $G_i^*$ that minimizes this bound given any acceptance criteria $\{f_i\}_{i=1}^L$, which coincides with $G_i^{\van}$ under the vanilla \ac{sd} setup with $f_i^{\van}$.

\subsubsection{Design of $G_{i}$.} We further characterize the property of the minimizer $\{G_i^*\}_{i=1}^L$ in the following proposition.

\begin{proposition}\label{prop:opt_resample}
    If $f_{i}(\cdot \,|\, x_{1:i},P,Q)\geq f_{i}^{\van}(\cdot \,|\, x_{1:i},P,Q)$ almost surely for all $x_{1:i}\in\calX^{i}$ and $i\in[L]$, the following holds almost surely
    \begin{align}
        G_i^{*}(\cdot \,|\, x_{1:i-1},P,Q\big)=
        G_i^{\van}(\cdot \,|\, x_{1:i-1},P,Q\big). \label{eq:G_design}
    \end{align}
\end{proposition}
The proof is provided in Appendix~\ref{app:resample_proof}. This result reveals a new property of $G_i^{\van}(\cdot \,|\, x_{1:i-1},P,Q\big)$ that it minimizes the TV upper bound when $f_{i}(\cdot \,|\, x_{1:i},P,Q)\geq f_{i}^{\van}(\cdot \,|\, x_{1:i},P,Q)$. This also provides a simple and explicit expression of the resampling distribution
when the acceptance probability is unchanged, i.e., identical to that in vanilla \ac{sd}. Experimental results corroborating this observation are presented in Section~\ref{sec:ablation}.
\begin{table*}[t]
\centering
\scalebox{0.93}{
\begin{tabular}{c|ccccccc}
\toprule
\textbf{Target Model} & \textbf{Method} & \textbf{CLIP (↑)} & \textbf{FID (↓)} & \textbf{IR (↑)} & \textbf{Acc. Len. (↑)} & \textbf{Latency/s (↓)} & \textbf{Speed-up/× (↑)} \\
\midrule
\multirow{4}{*}{{Lumina-mGPT}} 
& Target Model & 0.3330 & 28.99 & 0.6855 & 1.00 {\scriptsize($\pm$0.00)} & 170.14 {\scriptsize($\pm$1.32)} & 1.00 \\
& Eagle-1 & 0.3330 & 29.05 & 0.6883 & 2.76 {\scriptsize($\pm$0.07)} & 71.66 {\scriptsize($\pm$1.80)} & 2.37 \\ \cmidrule(lr){2-8}
& LANTERN++ ($\lambda$=2, $k$=10) & \textbf{0.3328} & 30.31 & 0.6697 & 2.99 {\scriptsize($\pm$0.07)} & 68.64 {\scriptsize($\pm$1.86)} & 2.48 \\
& \ourmethod{} ($\delta$=1.1) & 0.3325 & \textbf{30.30} & \textbf{0.6699} & \textbf{3.11} {\scriptsize($\pm$0.07)} & \textbf{63.24} {\scriptsize($\pm$1.83)} & \textbf{2.69} \\
\midrule
\multirow{5}{*}{{LlamaGen-XL}} 
& Target Model & 0.3162 & 21.08 & -0.0763 & 1.00 {\scriptsize($\pm$0.00)} & 10.11 {\scriptsize($\pm$0.82)} & 1.00 \\
& Eagle-1 & 0.3157 & 20.97 & -0.0859 & 2.42 {\scriptsize($\pm$0.15)} & 4.99 {\scriptsize($\pm$0.34)} & 2.03 \\ \cmidrule(lr){2-8}
& LANTERN++ ($\lambda$=2, $k$=10) & 0.3157 & 21.17 & -0.1155 & 2.67 {\scriptsize($\pm$0.18)} & 4.70 {\scriptsize($\pm$0.38)} & 2.15 \\
& \ourmethod{} ($\delta$=1.1) & \textbf{0.3167} & \textbf{21.02} & \textbf{-0.0997} & 2.73 {\scriptsize($\pm$0.16)} & 4.46 {\scriptsize($\pm$0.34)} & 2.27 \\
& \ourmethod{} ($\delta$=2) & 0.3154 & 21.20 & -0.1353 & \textbf{3.34} {\scriptsize($\pm$0.20)} & \textbf{3.72} {\scriptsize($\pm$0.27)} & \textbf{2.72} \\
\bottomrule
\end{tabular}
}
\caption{{Quantitative results with Lumina-mGPT and LlamaGen-XL as the target models.} The {bold} indicates the best results among the \ac{sd} methods that consider relaxations. \ourmethod{} achieves a better trade-off between latency and generation quality. The parameters $\lambda$ and $k$ are relaxation hyperparameters used in LANTERN++; we adopt the same values as in the experiments reported by~\citet{park2025lantern++}. Standard deviations are shown in parentheses. }\label{tab:main}
\label{tab:main-result}
\end{table*}

\subsubsection{Design of $f_{i}$.}

For clarity and conciseness, we now specify a concrete parameterization of the acceptance functions. We emphasize that the annealing property we present still holds for general choices of $f_i$, as formally shown in Appendix~\ref{app:formal_proof}. Since the acceptance threshold functions ${f_{i}(\tilx_{1:i}, P, Q)}_{i=1}^{L}$ can take various forms, we adopt a simple yet effective instantiation by introducing a multiplicative relaxation parameter $\omega_i$ into their formulation:
\begin{align}
    f_{i}(\tilx_{1:i}, P, Q; \omega_{i})\! = \min\!\left\{1, \frac{\omega_{i}\cdot P(\tilx_{i}\,|\,\pt, \tilx_{1:i-1})}{Q(\tilx_{i}\,|\,\pt, \tilx_{1:i-1})}\right\}\!.
    \label{eq:relax}
\end{align}
A straightforward choice is to set 
\begin{align}
    \text{(\unirelax)}\qquad\quad\omega_i = \delta\text{ for all $i \in [L]$,}\quad\qquad\quad
\end{align}
which we refer to as Uniform Relaxation of Speculative Decoding~(\unirelax{}) in the following, and we refer to $\delta$ as the relaxation budget. 

To further improve the efficiency of relaxed \ac{sd}, we conduct a perturbation analysis on the acceptance criterion $f_{i}$, to shed light on our choices of parameter $\omega_{i}$ other than a fixed schedule. 
In the following, we consider the perturbed $\{f_i\}_{i=1}^{L}$ when $L=2$ as
\begin{align*}
    \tilf_{1}(x_1, c_1, P,Q;\omega_{1})&=f_1(x_1,P,Q;\omega_{1})+c_1, \\
    \tilf_{2}((x_{1},x_{2}),c_2, P,Q;\omega_{2})&=f_2((x_{1},x_{2}),P,Q;\omega_{2})+c_2.
\end{align*}
Here we require that $|c_1|,|c_2|=o(1)$, i.e., the amplitudes of the perturbations are small as the common perturbation analysis~\citep{bonnans2013perturbation}. We denote the total variation bound, i.e., the right-hand side of Eqn.~\eqref{ieq:tv_token} under $\{\tilf_i\}_{i=1}^{2}$ and the corresponding $\{\tilG_i^{*}\}_{i=1}^{2}$ as $\texttt{TVB}(c_1,c_2)$.
\begin{proposition}[Annealing Property of the Relaxation Criterion]\label{prop:two_token_informal}
    Maintaining the same expectation of the number of accepted tokens for $\{\tilf_i\}_{i=1}^{2}$ and $\{f_i\}_{i=1}^{2}$ requires that
    \begin{align}
        c_1 = \frac{-\bbE_{X_1\sim Q}\big[f_1(X_1,P,Q;\omega_{i})\big]\cdot c_2}{1+\bbE_{(X_1,X_2)\sim Q}\Big[f_2\big((X_1,X_2),P,Q;\omega_{i}\big)\Big]+c_2}\label{eq:c1c2_stat_1}.
    \end{align}
    If some regularties conditions hold and $f_2^{\van}((x_1,x_2),P,Q)\leq f_2((x_1,x_2),P,Q;\omega_{i})$ pointwisely, we have the following for two perturbations $(c_1,c_2)$ and $(\tilc_1,\tilc_2)$ satisfying Eqn.~\eqref{eq:c1c2_stat_1} but with different signs, i.e., $c_1,\tilc_2 >0$, $c_2,\tilc_1 <0$,
    \begin{align}
        \texttt{TVB}(c_1,c_2)<\texttt{TVB}(\tilc_1,\tilc_2).\label{ieq:c1c2}
    \end{align}
\end{proposition}
The mentioned regularity conditions and the formal statement of this proposition are provided in Appendix~\ref{app:formal_proof}. This proposition indicates that increasing relaxation at an earlier position while reducing it at a later one—as specified in~\eqref{eq:c1c2_stat_1}—leads to lower distributional bias than the reverse, i.e., \eqref{ieq:c1c2}. When $L > 2$, a similar conclusion can be drawn by comparing each pair of neighboring positions. 

Therefore, we propose an \emph{annealed relaxation of speculative decoding} ({\textbf{\ourmethod}}) scheme for the acceptance criterion (see Algorithm~\ref{alg:approx_SD}). Instead of using a fixed relaxation parameter across all drafted positions, our approach progressively tightens the acceptance condition as decoding proceeds, i.e., $\omega_{0}>\omega_{1}>\ldots>\omega_{L}$. 
Specifically, we define the $\omega_{i}$ in Eqn.~\eqref{eq:relax} to be
\begin{align}
    \text{(\ourmethod{})}\quad\quad \omega_i = \delta \cdot \exp(-\nu\cdot i-\mu), \qquad\qquad
    \label{eq:sdar}
\end{align}
for all $i \in [L]$
with the decaying coefficient $\nu > 0$
and normalizing parameter $\mu$ such that $\sum_{i=1}^{L} \exp(-\nu\cdot i-\mu) = L$. \ourmethod{} is expected to induce less  shift than \unirelax{} under the same accepted length. In Section~\ref{sec:ablation}, we show that empirically \ourmethod{} achieves better image quality than \unirelax{} at the same inference speed.

\section{Experiments}\label{sec:exp}

In this section, we evaluate our method through quantitative and qualitative results, comparing it with vanilla \ac{sd} and LANTERN++. We also show visual results under different relaxation budgets $\delta$ to show  the latency-quality trade-off.

\subsection{Implementation Details}

We conduct extensive experiments to assess the effectiveness of \ourmethod{} on LlamaGen-XL~\citep{sun2024autoregressive} (775M) and Lumina-mGPT~\citep{liu2024lumina} (7B) for accelerating visual AR inference, using a single NVIDIA A100-SXM4-40GB GPU. See Appendix~\ref{app:imple-detail} for more details.

\subsubsection{Draft Model Training and Inference}
For training the draft models, we follow a pipeline similar to Eagle~\citep{li2024eagle} and LANTERN++~\citep{park2025lantern++}. We evaluate all methods on 5k randomly selected captions from the MS-COCO 2017 validation set~\citep{lin2014microsoft}, following the inference setup of LANTERN~\citep{jang2024lantern} and LANTERN++~\citep{park2025lantern++}.

\subsubsection{Metrics}
We measure acceleration by average accepted length and latency, where the former is the ratio of generated tokens to speculative decoding rounds.
Generation quality is evaluated using the Fréchet Inception Distance (FID)~\citep{heusel2017gans}, CLIP score~\citep{hessel2021clipscore}, and ImageReward (IR)~\citep{xu2023imagereward}. 

\subsubsection{Baselines} 
We benchmark our method against Eagle-1~\citep{li2024eagle}, which serves as a baseline for lossless speculative decoding, and LANTERN++~\citep{park2025lantern++}, a state-of-the-art relaxation-based speculative decoding approach. We adopt the static tree structure in Eagle-1~\citep{li2024eagle}, which has been demonstrated by \citet{park2025lantern++} to outperform the dynamic tree structure~\citep{li2024eagle2}.

\subsection{Experimental Results}

\begin{figure*}[t]
    \centering
    \subfigure[FID v.s. Accepted Length on \newline LlamaGen-XL. ]{\includegraphics[width=0.237\textwidth]{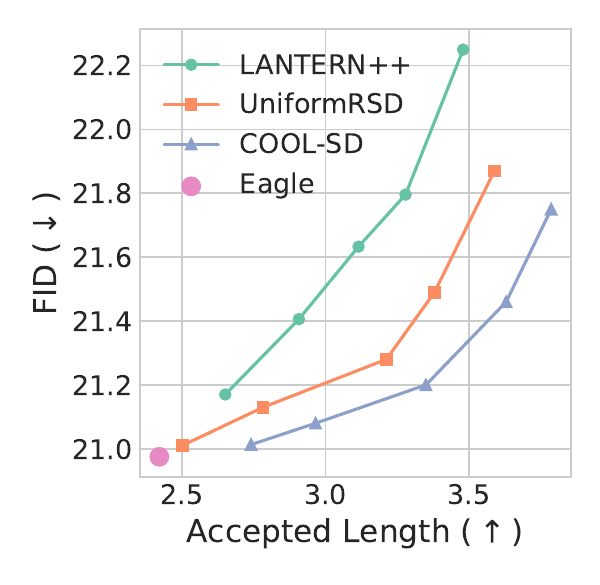}} \;
    \subfigure[FID v.s. Latency on  \newline LlamaGen-XL. ]{\includegraphics[width=0.237\textwidth]{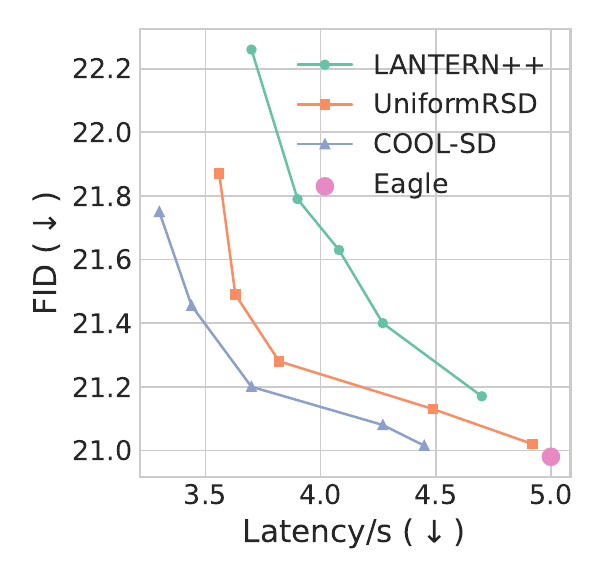}} \;
    \subfigure[FID v.s. Accepted Length on Lumina-mGPT. ]{\includegraphics[width=0.237\textwidth]{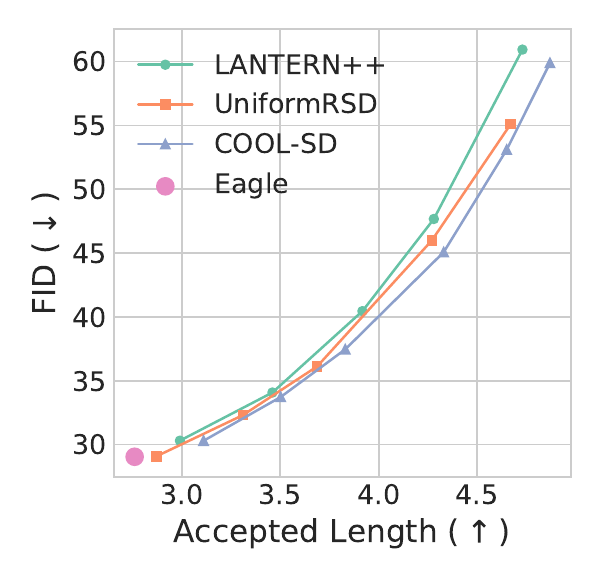}} \;
    \subfigure[FID v.s. Latency on \newline Lumina-mGPT. ]{\includegraphics[width=0.237\textwidth]{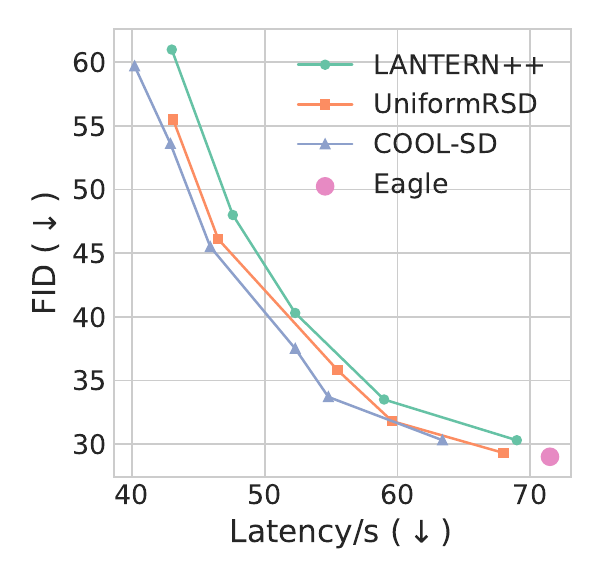}}
    \caption{{The trade-off curves between imaging quality (evaluated by FID) and accepted length/lantency. } We compared \ourmethod{} with \unirelax{} and LANTERN++ with $k=10$ on two target models: LlamaGen-XL and Lumina-mGPT. We tested on $5000$ random sampled captions from MSCOCO validation set.  }\label{fig:abl-delta}
\end{figure*}

\begin{figure}
    \centering
    \includegraphics[width=\linewidth]{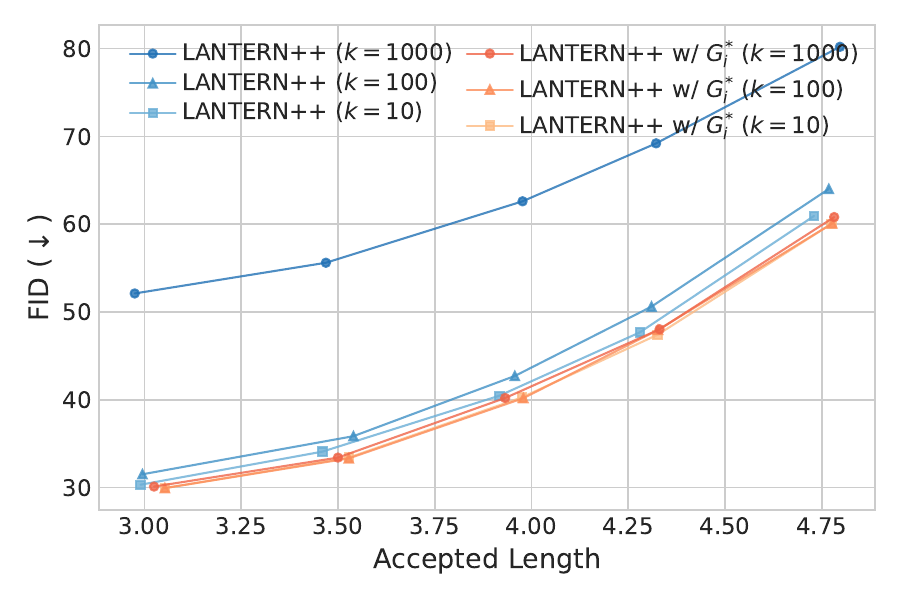}
    \caption{{Ablation study on LANTERN++ with our resampling distribution $G_{i}^{*}$},  under different settings of $k$. }
    \label{fig:lantern_abl}
\end{figure}

\subsubsection{Quantitative Results}

Table~\ref{tab:main} summarizes our main results. Compared to the lossless Eagle-1, \ourmethod{} achieves significantly faster inference with only minor quality degradation. The speed-quality trade-off in \ourmethod{} is tunable via the relaxation budget $\delta$: smaller $\delta$ maintains near-baseline quality with faster inference, while larger $\delta$ offers greater speedups with modest quality loss.

Specifically,
on LlamaGen, our method increases the average number of accepted steps from $2.42$ to $2.73$ with negligible loss in FID and CLIP Score. When the accepted length is further raised to $3.34$, the image quality remains nearly unchanged. Similarly, on Lumina-mGPT, we observe consistent improvements: the accepted length rises from $2.75$ to $3.15$, leading to a reduction of approximately 10 seconds per image on the latency, while the FID and CLIP Score degradation remains within an acceptable range.

Compared to the relaxation-based \ac{sd} method LANTERN++, \ourmethod{} consistently produces higher-quality images under the same latency constraints or achieves greater acceleration at similar quality, showing  superior efficiency. For a fair comparison with the hyperparameters $\lambda=2, k=10$ used in the LANTERN++ paper, we set the hyperparameter $\delta$ of \ourmethod{} to $1.1$ or $2.0$, and $\nu$ to $0.7$. As shown in Figure~\ref{fig:abl-delta} of the next section, even when varying $\delta$ for \ourmethod{} and $\lambda, k$ for LANTERN++, \ourmethod{} still outperforms LANTERN++ across settings. Additional comparisons with Speculative Jacobi Decoding (SJD)~\citep{teng2024accelerating} are presented in Appendix~\ref{app:imple-detail}.

\subsubsection{Qualitative Results}

In Figure~\ref{fig:visual}, we present visualizations of our method under different choices of relaxation budget $\delta$ from $1.1$ to $3.0$. 
Increasing $\delta$ effectively improves the inference speed, demonstrating that the trade-off between speed and generation quality can be flexibly controlled through this parameter. 
As shown in the figure, the visual quality remains comparable to vanilla \ac{sd} even as the inference speed increases from $\times 2.37$ to $\times 3.10$. Although further increasing the inference speed to $\times 3.70$ and beyond may result in some degradation in image quality, this trade-off remains easily manageable and can be tuned to meet practical needs.

We also emphasize that our method achieves superior speed-quality trade-off compared to LANTERN++~\citep{park2025lantern++}.  The corresponding quantitative results are presented in Section~\ref{sec:ablation}, while qualitative comparisons with LANTERN++ are provided in Appendix~\ref{app:imple-detail}. 

\section{Ablation Study}\label{sec:ablation}
This section focuses on ablation studies across a wide range of parameter settings for \ourmethod{}, comparing it with LANTERN++. We also analyze the impact of our resampling distribution when applied to LANTERN++.

\subsection{Trade-off Between Speed and Quality}
In this paper, we propose a principled approach \ourmethod{} to 
control the trade-off between inference speed and image quality by adjusting $\delta$ in the range from $1.1$ to $4.0$, and $\nu = 0.7$ in \ourmethod{}. 
Figure~\ref{fig:abl-delta} illustrates the relationship between FID and accepted length/latency for the two target models, LlamaGen-XL~\citep{sun2024autoregressive} and Lumina-mGPT~\citep{liu2024lumina}. Evaluating different methods requires examining the trade-off between generation quality and acceleration. As shown in Figures~\ref{fig:abl-delta}(a) and (b), for the same FID level (i.e., along a horizontal slice of the plot), \ourmethod{} achieves a greater accepted length and lower latency. Conversely, at a fixed acceptance length or latency, \ourmethod{} attains a lower FID compared to LANTERN++. Moreover, due to its simple implementation—free from the nearest-token probability retrieval and summation required by LANTERN++—\ourmethod{} achieves additional latency improvements, a decisive factor for real-world applications.

We further compare \ourmethod{} with \unirelax{}, a straightforward setting in which all drafting steps share the same relaxation parameter, as shown in Figure~\ref{fig:abl-delta}. The results demonstrate that the annealing strategy significantly enhances relaxation performance, yielding 
better trade-off between generation quality and inference speed.

\subsection{Principled Resampling Strategy}

LANTERN~\citep{jang2024lantern} and LANTERN++~\citep{park2025lantern++} proposed to perform relaxation on \ac{sd} based on image token similarity. However, a mathematically rigorous proof is missing for their corresponding resampling distribution. To illustrate the importance of the resampling distribution and the effectiveness of our method, we conduct an experiment on LANTERN++ and LANTERN++ with our resampling distribution~$G^*$ in Eqn.~\eqref{eq:opt_G}. Our experiments are conducted under different settings of $k$—the hyperparameter in LANTERN++ that controls the number of nearest tokens in the latent space considered as similar. 
The result is shown in Figure~\ref{fig:lantern_abl}. 
As demonstrated, with our resampling distribution, LANTERN++'s performance is improved regardless of the settings of hyperparameter $k$. 
Also, since LANTERN++ adopts an 
inferior resampling strategy, as $k$ is set larger, the bias introduced by their algorithm grows. This issue is empirically mitigated by our resampling distribution, as evidenced by the orange lines in Figure~\ref{fig:lantern_abl}.

\section{Conclusion}
In this paper, we introduce \ourmethod{}, a principled relaxed \ac{sd} framework that consistently improves the speed–quality trade-off in auto-regressive image generation through two key insights: (1) establishing a tight bound on distributional bias and deriving token-wise optimal resampling, and (2) developing a provably effective decaying acceptance schedule. Extensive experiments demonstrate that \ourmethod{} outperforms existing speculative decoding variants in both efficiency and fidelity.

\section*{Acknowledgments}
Xingyao Li  acknowledges the support of Sea AI Lab for providing computational resources. This work is also supported by the Singapore Ministry of Education Academic Research Fund (AcRF) Tier 1 Grant (No. A-8000981-00-00).

\bibliography{aaai2026}

\newpage
$ $
\newpage
\appendix
\section{Notations} \label{app:notation}

Let $[N]:=\{1,\cdots,N\}$. For any finite or infinite set $\calZ$, we denote the uniform distribution on it as $\unif(\calZ)$. For a finite set $\calX$, we denote the set of all the distributions on it as $\calP(\calX)=\{P:\calX\rightarrow [0,1] \,|\, \sum_{x\in\calX}P(x)=1, P(x)\geq 0 \text{ for all }x\in\calX\}$. The closure of it is denoted as $\bar{\calP}(\calX)=\{P:\calX\rightarrow [0,1] \,|\, \sum_{x\in\calX}P(x)\leq 1, P(x)\geq 0 \text{ for all }x\in\calX\}$. The space of all the sequences whose component is in $\calX$ is denoted as $\calX^{*}$, and we use $x_{1:L}\in\calX^{L}\subseteq\calX^{*}$ to denote a $L$-length sequence. 
The set of transition kernels on $\calX^{*}$ is denoted as $\calT(\calX)=\{P:\calX\times\calX^{*}\rightarrow [0,1]\,|\, P(\cdot |x_{1:L})\in\calP(\calX)\text{ for all }x_{1:L}\in\calX^{*}\}$. For any function $f:\calX\rightarrow\bbR_{+}$ with $\sum_{x\in\calX}f(x)>0$, we define the distribution induced by it as $\nor(f(\cdot))=f(\cdot)/\sum_{x^{\prime}\in\calX}f(x^{\prime})$. 
We denote the positive part of a function $f$ as $[f(\cdot)]_{+}=\max\{0,f(\cdot)\}$. The random variable and its realization are respectively denoted as the uppercase and lowercase letters, i.e., $X=x$. For two distributions $P,Q\in\calP(\calX)$, the total variation distance between them is $\tv(P,Q)=\max_{\calA\subseteq \calX}P(\calA)-Q(\calA)$. In this paper, we define $0/0=1$.
\section{Additional Related Works}\label{app:related_works}

Beyond the standard \ac{sd} framework, recent works extend \ac{sd} from an algorithmic perspective, such as adaptive draft selection~\citep{hou2025banditspec} and efficient lossless verification for long-context generation~\citep{yang2025longspec}.

In \ac{ar} image generation, alongside \ac{sd}, several works focus on improving efficiency. We highlight that the effectiveness of these methods can be further enhanced by integrating and refining \ac{sd} within their respective frameworks. One line of work exploits spatial locality to enable the simultaneous generation of multiple tokens. For instance, ZipAR~\citep{hezipar} generates image tokens based on a zip-rank ordering, allowing parallel generation along diagonals. \citet{wang2025parallelized} proposes generating tokens from spatially distant pixel positions in parallel, while \citet{pang2024next} adopts a patch-wise strategy to generate multiple nearby tokens in a single \ac{ar} step. Distinct from these methods, \citet{teng2024accelerating} incorporate \ac{sd} into Jacobi decoding to further accelerate generation.

\section{Additional Experiments}\label{app:imple-detail}

\subsection{Evaluation Metrics}
Fr\'echet Inception Distance (FID)~\citep{heusel2017gans} measures the distributional similarity between generated and real images in the feature space of a pretrained Inception network. A lower FID indicates that the generated images are more similar to real ones in terms of visual statistics and diversity. We employ Inception-V3~\citep{szegedy2016rethinking} to extract image features for FID computation. 

CLIP Score~\citep{hessel2021clipscore} evaluates the semantic alignment between an image and its corresponding text prompt by computing the cosine similarity between their embeddings extracted from the CLIP model. A higher CLIP score reflects better text–image consistency. We use OpenAI’s ViT-B/32 model~\citep{radford2021learning} for embedding extraction. 

ImageReward~\citep{xu2023imagereward} is further adopted to assess human preference for the generated images, providing a reward signal aligned with human aesthetic and semantic judgments. 

\subsection{Additional Implementation Details}
The draft model for LlamaGen-XL is trained on 100k image–caption pairs randomly sampled from Laion-COCO~\citep{laioncoco2022}, while for Lumina-mGPT, 30k images are generated using Lumina-mGPT from captions randomly sampled from the MS-COCO 2017 training set~\citep{lin2014microsoft}.

During the draft model training, each sample has a  probability of $0.1$ of being randomly converted into a null-conditioned sample. During inference, the draft model generates predictions separately for conditional and unconditional inputs, and the final speculation is obtained using classifier-free guidance~\citep{ho2022classifier}. 

In both the qualitative and quantitative results shown in the paper, we use seeds 42-46. 

\subsection{Ablation Study on the Annealing Schedule}

\begin{figure}[t]
    \centering
    \subfigure[FID v.s. Accepted Length on \newline LlamaGen-XL. ]{\includegraphics[width=0.23\textwidth]{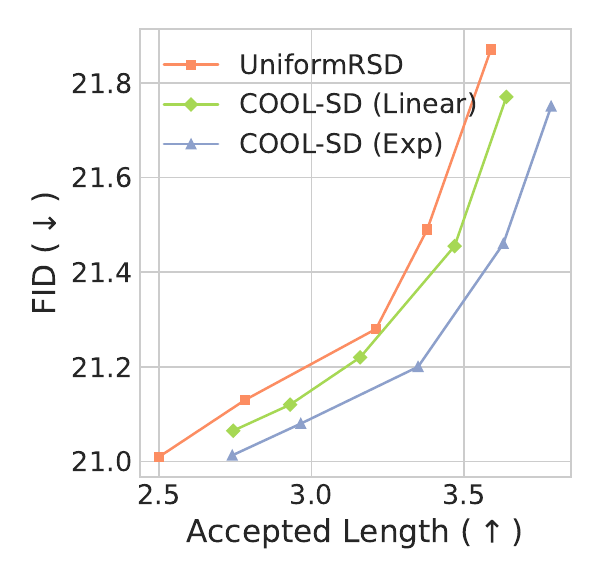}} 
    \subfigure[FID v.s. Accepted Length on \newline Lumina-mGPT. ]{\includegraphics[width=0.23\textwidth]{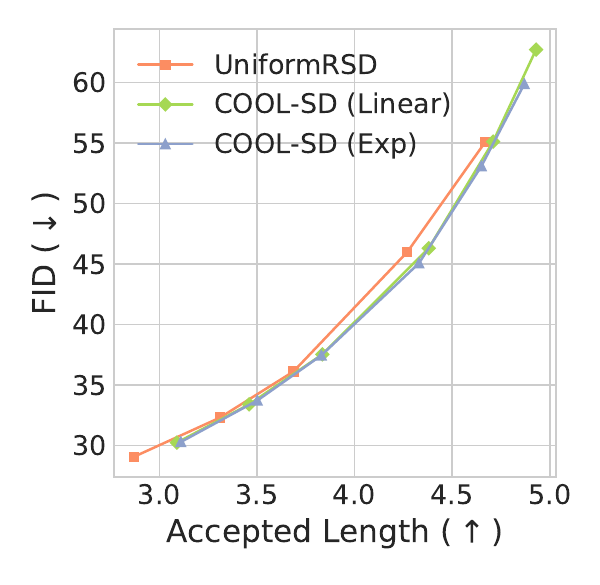}} 
    \caption{{Ablation study on the effects of the annealing schedule on trade-off curves.} We compare \ourmethod{} using a linear or exponential annealing schedule against the uniform schedule \unirelax{}, across two target models: LlamaGen-XL and Lumina-mGPT. Evaluations are conducted on $5,000$ randomly sampled captions from the MS-COCO validation set.}\label{fig:abl-schedule}
\end{figure}

As suggested by Proposition~\ref{prop:two_token_informal}, a decaying acceptance schedule helps reduce distributional shift at a fixed accepted length. We implement \ourmethod{} using both a linear and an exponential decay schedule. The exponential schedule is defined in Eqn.~\ref{eq:sdar}, while the linear schedule is given by
\begin{align}
    \omega_{i} = \delta\cdot L\cdot\frac{\tilomega_{i}}{\sum_{i=1}^{L}\tilomega_{i}}, \quad \text{where}\quad
    \tilomega_{i} = \frac{\ell - i}{\ell(\ell+1)}. 
\end{align}
Here, $\delta$ denotes the relaxation budget, and $\ell$ is a hyperparameter controlling the slope of the linear decay. In our experiments, we set $\ell = 8$.

Experimental results are shown in Figure~\ref{fig:abl-schedule}. We evaluate both variants on LlamaGen-XL~\citep{sun2024autoregressive} and Lumina-mGPT~\citep{liu2024lumina}. On LlamaGen-XL, \ourmethod{} with the exponential schedule (Exp) achieves a better FID–Accepted Length trade-off than its linear counterpart (Linear), and both outperform the uniform schedule \unirelax{}. On Lumina-mGPT, \ourmethod{} (Linear) performs comparably to \ourmethod{} (Exp) and again surpasses \unirelax{}. These observations align well with our theoretical insights.

\subsection{Significance Analysis on Quantitative Experiment}
On Lumina-mGPT~\citep{liu2024lumina}, using the settings in Table~\ref{tab:main-result}, \ourmethod{} yields a statistically significant increase in acceptance length over LANTERN++ ($t = 18.11$, $p < 10^{-60}$) across $1000$ samples, while maintaining a comparable FID~\citep{heusel2017gans}, demonstrating the effectiveness of our approach. Similarly, on LlamaGen-XL~\citep{sun2024autoregressive}, \ourmethod{} with $\delta = 1.1$ achieves a significantly higher average accepted length than LANTERN++ ($t = 7.31$, $p < 10^{-12}$).

\subsection{Comparison with Speculative Jacobi Decoding}
Speculative Jacobi Decoding (SJD)~\citep{teng2024accelerating} is a speculative decoding approach designed to accelerate auto-regressive image generation, leveraging Jacobi updates to better adapt speculative decoding to image domain. We compare \ourmethod{} with SJD in terms of efficiency and quality in Table~\ref{tab:jacobi}. While SJD serves as a lossless acceleration method, \ourmethod{} achieves substantially higher speed-up ratios with only a slight reduction in image quality. 

\begin{table*}[t]
\centering
\scalebox{1}{
\begin{tabular}{c|cccccc}
\toprule
\textbf{Target Model} & \textbf{Method} & \textbf{CLIP (↑)} & \textbf{FID (↓)} & \textbf{ImageReward (↑)} & \textbf{Speed-up/× (↑)} \\
\midrule
\multirow{3}{*}{{Lumina-mGPT}} 
& Target Model & 0.3330 & 28.99 & 0.6855 & 1.00 \\
& SJD & 0.3305 & 31.81 & 0.6764 & 2.06 \\
& \ourmethod{} ($\delta$=1.1) & 0.3325 & 30.30 & 0.6699 & 2.69 \\
\midrule
\multirow{4}{*}{{LlamaGen-XL}} 
& Target Model & 0.3162 & 21.08 & -0.0763 & 1.00 \\
& SJD & 0.3160 & 21.01 & -0.1006 & 1.84 \\
& \ourmethod{} ($\delta$=1.1) & 0.3167 & 21.02 & -0.0997 & 2.27 \\
& \ourmethod{} ($\delta$=2) & 0.3154 & 21.20 & -0.1353 & 2.72 \\
\bottomrule
\end{tabular}
}
\caption{{Quantitative comparison against Speculative Jacobi Decoding (SJD).}}
\label{tab:jacobi}
\end{table*}

\subsection{Qualitative Comparison with LANTERN++}

\begin{figure*}[t]
    \centering
    \includegraphics[width=0.9\textwidth]{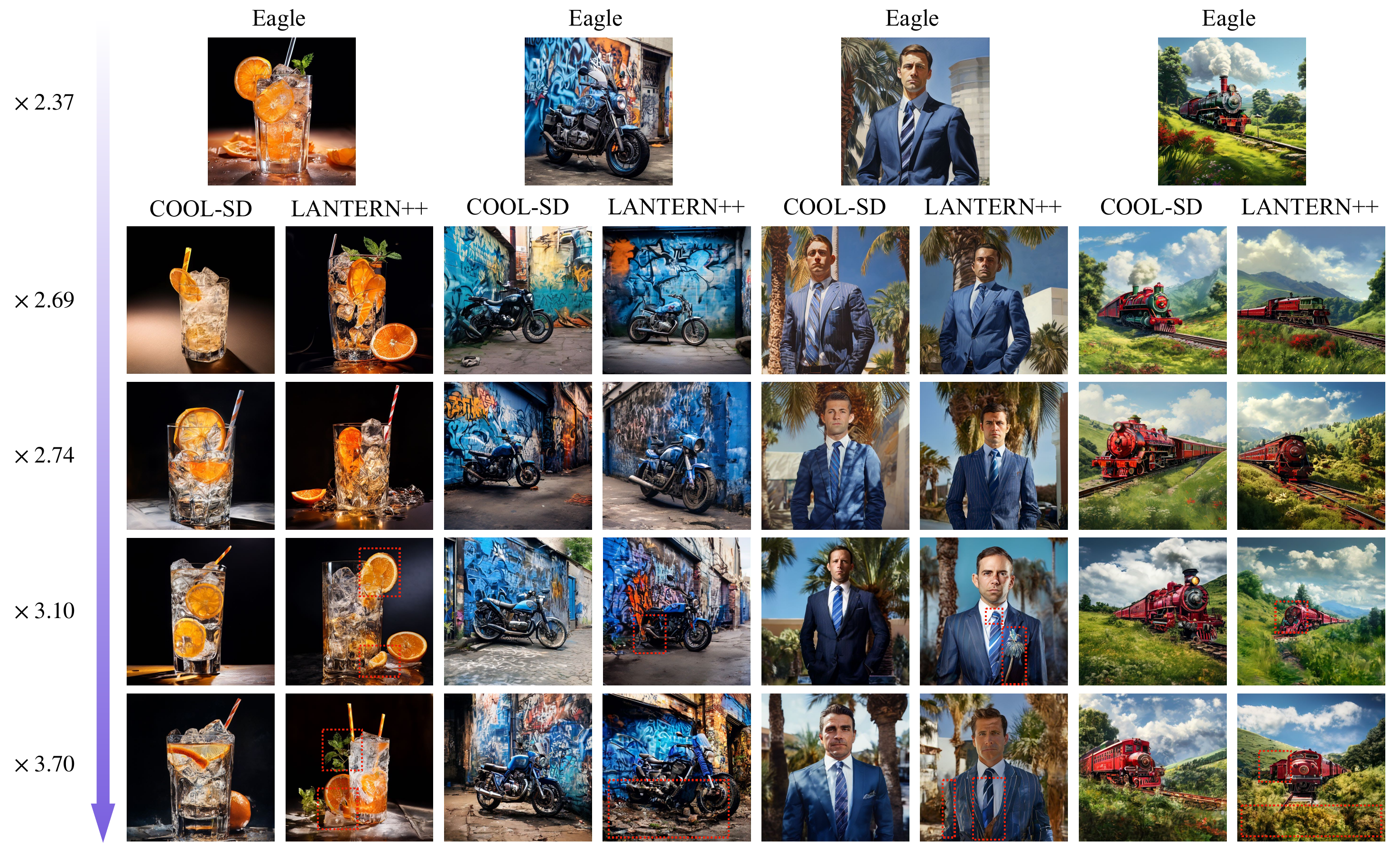}
    \caption{{Qualitative comparison of \ourmethod{} and LANTERN++}. We compare outputs from \ourmethod{} and LANTERN++ on Lumina-mGPT under different speed-up. The speed-up factor is annotated to the left of each row. The first row shows images generated by Eagle-1 without any relaxation, serving as the baseline. Visible artifacts are highlighted with red boxes. }\label{fig:app_visual}
\end{figure*}

We present qualitative comparisons between \ourmethod{} and LANTERN++~\citep{park2025lantern++} in Figure~\ref{fig:app_visual} under similar speed-up factors. Both methods exhibit a gradual degradation in image quality as the speed-up factor increases, owing to the relaxation of the acceptance criterion. However, LANTERN++ suffers from noticeably stronger degradation than \ourmethod{}. \ourmethod{} preserves image fidelity comparable to the lossless baseline up to a $\times 3.10$ speed-up, whereas LANTERN++ begins to show visible artifacts and patchy backgrounds at this point, which become increasingly severe when the speed-up reaches $\times 3.70$.

\section{Proof of Theorem~\ref{thm:t_tvbd}}\label{app:proof_main}
The proof of Theorem~\ref{thm:t_tvbd} takes three steps:
\begin{itemize}
    \item Derive the expression of the token distribution $\hatP$.
    \item Bound the total variation between $\hatP$ and $P$.
    \item Derive the expression of the number of accepted tokens.
\end{itemize}

\textbf{Step 1: Derive the expression of the token distribution $\hatP$.}

For the conciseness of notations, we will omit the prefix $\pt$ in the proof. We will adopt $\bbP$ to denote the probability induced by Algorithm~\ref{algo:t_asd} and the virtual generation process. From the definition of $\hatP$ and the Algorithm~\ref{algo:t_asd}, we have that
\begin{align}
    \hatP(X_{1:L+1}=x_{1:L+1}) &= P(X_{L+1}=x_{L+1}\mid X_{1:L}=x_{1:L})
    \\*
    &\qquad
    \cdot\hatP(X_{1:L}=x_{1:L}),\label{eq:condp}
\end{align}
where the equality results from the virtual generation procedure and the fact that $\tau\leq L$. Then we only need to focus on $\hatP(X_{1:L}=x_{1:L})$. We will represent its expression in two different ways: its explicit expression and its recursive relationship. We first derive the recursive relationship. For any $i\in[L]$, we have that
\begin{align}
    &\hatP(X_{1:i}=x_{1:i})\nonumber\\
    &\quad= \sum_{\tilx_{1:i}\in\calX^{i}}\sum_{j=0}^{i-1}Q(\tilX_{1:i}=\tilx_{1:i})\cdot\bbP(\tau=j\mid  \tilX_{1:i}=\tilx_{1:i})
    \\
        &
        \hphantom{\quad= \sum_{\tilx_{1:i}\in\calX^{i}}\sum_{j=0}^{i-1}}
        \cdot\bbP(X_{1:i}=x_{1:i}\mid \tau=j, \tilX_{1:i}=\tilx_{1:i})\nonumber\\
    &\quad\qquad+\sum_{\tilx_{1:i}\in\calX^{i}} Q(\tilX_{1:i}=\tilx_{1:i})\cdot\bbP(\tau\geq i\mid  \tilX_{1:i}=\tilx_{1:i})
    \\&
    \hphantom{\quad\qquad+\sum_{\tilx_{1:i}\in\calX^{i}}}
    \cdot\bbP(X_{1:i}=x_{1:i}\mid \tau\geq i, \tilX_{1:i}=\tilx_{1:i}).\label{eq:full_prob}
\end{align}
For ease of notation, we abbreviate $X_{1:l}=x_{1:l}$ and $\tilX_{1:l}=\tilx_{1:l}$ as $x_{1:l}$ and $\tilx_{1:l}$ for any $l\in[L+1]$, respectively. For the first term in the right-hand side of Eqn.~\eqref{eq:full_prob}, we have
\begin{align}
    &\sum_{\tilx_{1:i}\in\calX^{i}} Q(\tilX_{1:i}=\tilx_{1:i})
    \cdot\bbP(\tau=j\mid  \tilX_{1:i}=\tilx_{1:i})
    \\&
    \hphantom{\sum_{\tilx_{1:i}\in\calX^{i}}}
    \cdot \bbP(X_{1:i}=x_{1:i}\mid \tau=j, \tilX_{1:i}=\tilx_{1:i})\nonumber\\
    &\quad = \sum_{\tilx_{1:i}\in\calX^{i}}Q(\tilx_{1:i})\cdot \big(1-f_{j+1}(\tilx_{1:j+1},P,Q)\big)
    \\&
    \hphantom{\quad = \sum_{\tilx_{1:i}\in\calX^{i}}}\cdot\prod_{k=1}^{j}f_{k}(\tilx_{1:k},P,Q) \cdot \prod_{k=1}^{j} \bbI\{x_k=\tilx_k\} \nonumber\\
    &\quad\qquad\qquad \cdot G_{j+1}(x_{j+1}\mid \tilx_{1:j}, P, Q)\cdot P(x_{j+2:i}\mid  x_{1:j+1})\nonumber\\
    &\quad = \sum_{\tilx_{j+1}\in\calX}Q(x_{1:j})\cdot Q(\tilx_{j+1}\mid x_{1:j})
    \\
    &
    \hphantom{\quad = \sum_{\tilx_{j+1}\in\calX}}\cdot
    \Big(1-f_{j+1}\big((x_{1:j},\tilx_{j+1}),P,Q\big)\Big)
    \\&
    \hphantom{\quad = \sum_{\tilx_{j+1}\in\calX}}
    \cdot\prod_{k=1}^{j}f_{k}(x_{1:k},P,Q)\cdot  G_{j+1}(x_{j+1}\mid  x_{1:j},P,Q)\nonumber\\
    &\quad\qquad\qquad \cdot P(x_{j+2:i}\mid  x_{1:j+1})\nonumber\\
    &\quad = Q(x_{1:j})\cdot G_{j+1}(x_{j+1}\mid  x_{1:j},P,Q)\cdot P(x_{j+2:i}\mid  x_{1:j+1})\nonumber\\
    &\quad\qquad\qquad \cdot  \sum_{\tilx_{j+1}\in\calX} Q(\tilx_{j+1}\mid x_{1:j})
    \\&
    \quad\qquad\qquad
    \cdot \Big(1-f_{j+1}\big((x_{1:j},\tilx_{j+1}),P,Q\big)\Big)
    \\&
    \quad\qquad\qquad
    \cdot\prod_{k=1}^{j}f_{k}(x_{1:k},P,Q)\nonumber\\
    &\quad = \barP(x_{1:L+1},j,G)\cdot \barr(x_{1:L+1},j,f),\label{eq:il-1}
\end{align}
where $\bbI\{\cdot\}$ is the indicator function, the first equality results from Algorithm~\ref{algo:t_asd}, the second and third equalities result from integrating and rearranging the terms, and we define the following quantities in the last equality.
\begin{align*}
    &\barP(x_{1:i},j,G) = Q(x_{1:j})\cdot G_{j+1}(x_{j+1}\mid  x_{1:j},P,Q)\\*
    &\qquad\cdot P(x_{j+2:i}\mid  x_{1:j+1}),\\
    &\barr(x_{1:i},j,f) =\sum_{\tilx_{j+1}\in\calX} Q(\tilx_{j+1}\mid x_{1:j})\\
    &\qquad \cdot\Big(1-f_{j+1}\big((x_{1:j},\tilx_{j+1}),P,Q\big)\Big)\cdot\prod_{k=1}^{j}f_{k}(x_{1:k},P,Q),
\end{align*}
where $j\in[i-1]$. For the second term in the right-hand side of Eqn.~\eqref{eq:full_prob}, we have that
\begin{align}
    &\sum_{\tilx_{1:i}\in\calX^{i}} Q(\tilX_{1:i}=\tilx_{1:i})\cdot\bbP(\tau\geq i\mid  \tilX_{1:i}=\tilx_{1:i})\\
    &\qquad\cdot\bbP(X_{1:i}=x_{1:i}\mid \tau\geq i, \tilX_{1:i}=\tilx_{1:i})\nonumber\\
    &\quad = \sum_{\tilx_{1:i}\in\calX^{i}} Q(\tilX_{1:i}=\tilx_{1:i})\\
    &\quad\qquad \cdot \prod_{k=1}^{i}f_{k}(\tilx_{1:k},P,Q)\prod_{k=1}^{i} \bbI\{x_{k}=\tilx_{k}\}\nonumber\\
    &\quad = Q(x_{1:i})\prod_{k=1}^{i}f_{k}(x_{1:k},P,Q),\label{eq:iL}
\end{align}
where the equalities result from Algorithm~\ref{algo:t_asd}. Combining Eqn.~\eqref{eq:full_prob}, \eqref{eq:il-1} and \eqref{eq:iL}, we have that
\begin{align}
    &\hatP(X_{1:i}=x_{1:i})=Q(x_{1:i})\prod_{k=1}^{i}f_{k}(x_{1:k},P,Q)\\
    &\hphantom{\hatP(X_{1:i}=x_{1:i})}\quad+\sum_{j=0}^{i-1}\barP(x_{1:i},j,G)\cdot \barr(x_{1:i},j,f).\label{eq:joint}
\end{align}
This equation implies a recursive expression of the marginal probability of $\hatP$ as
\begin{align}
    &\hatP(X_{1:i+1}=x_{1:i+1})\nonumber\\
    &\quad=Q(x_{1:i+1})\prod_{k=1}^{i+1}f_{k}(x_{1:k},P,Q)\\
    &\quad\qquad+\sum_{j=0}^{i}\barP(x_{1:i+1},j,G)\cdot \barr(x_{1:i+1},j,f)\nonumber\\
    &\quad = \barP(x_{1:i+1},i,G)\cdot \barr(x_{1:i+1},i,f)\\
    &\quad\qquad+\sum_{j=0}^{i-1}\barP(x_{1:i+1},j,G)\cdot \barr(x_{1:i+1},j,f)\\
    &\quad\qquad + Q(x_{1:i+1})\prod_{k=1}^{i+1}f_{k}(x_{1:k},P,Q)\nonumber\\
    &\quad = \bigg[\hatP(X_{1:i}=x_{1:i})-Q(x_{1:i})\prod_{k=1}^{i}f_{k}(x_{1:k},P,Q)\bigg]\\
    &\quad\qquad \cdot P(x_{i+1}\mid  x_{1:i})
    + \barP(x_{1:i+1},i,G)\cdot \barr(x_{1:i+1},i,f) \\
    &\quad\qquad +Q(x_{1:i+1})\prod_{k=1}^{i+1}f_{k}(x_{1:k},P,Q)\nonumber\\
    &\quad = \hatP(X_{1:i}=x_{1:i})\cdot P(x_{i+1}\mid  x_{1:i})
    \\*
    &\quad\qquad+ Q(x_{1:i})\prod_{k=1}^{i}f_{k}(x_{1:k},P,Q)\nonumber
    \\*
    &\quad\qquad 
    \cdot\! \bigg[G_{i+1}(x_{i+1}\mid x_{1:i},P,Q)
    \\*
    &\quad\qquad \quad
    \cdot\!\!\!\!\sum_{\tilx_{i+1}\in\calX}\!\!\Big[1\!-\!f_{i+1}\big((x_{1:i},\tilx_{i+1}),P,Q\big)\!\Big]Q(\tilx_{i+1}| x_{1:i})\nonumber\\*
    &\quad\quad +Q(x_{i+1}|x_{1:i}) f_{i+1}(x_{1:i+1},P,Q)\!-\!P(x_{i+1}| x_{1:i})\bigg],
    \label{eq:recur}
\end{align}
where the first and second equalities result from Eqn.~\eqref{eq:joint}, and the third and fourth equalities result from rearranging the terms. Based on this recursive expression, we can derive the explicit expression of the distribution $\hatP$ as 
\begin{align}
    &\hatP(X_{1:i}=x_{1:i})
    \\
    &\quad= \sum_{j=1}^{i-1}P(x_{j+2:i}\mid x_{1:j+1})Q(x_{1:j})\cdot\prod_{k=1}^{j}f_{k}(x_{1:k},P,Q)\\
    &\;\qquad\cdot \bigg[Q(x_{j+1}\mid x_{1:j})f_{j+1}(x_{j+1},P,Q)-P(x_{j+1}\mid x_{1:j})\\
    &\;\qquad +G_{j+1}(x_{j+1}\mid x_{1:j},P,Q)\\
    &\quad\qquad\cdot\!\!\!\sum_{\tilx_{j+1}\in\calX}\!\!\Big[1\!-\!f_{j+1}\big((x_{1:j},\tilx_{j+1}),P,Q\big)\Big]Q(\tilx_{j+1}| x_{1:j})\bigg]\nonumber\\
    &\;\qquad +P(x_{2:i}\mid x_{1})\bigg[Q(x_{1})f_{1}(x_{1},P,Q)\\
    &\;\qquad+G_{1}(x_{1}\mid  P,Q)\sum_{\tilx_{1}\in\calX}\Big[1-f_{1}(\tilx_{1},P,Q)\Big]Q(\tilx_{1})\bigg].\label{eq:joint1}
\end{align}

\textbf{Step 2: Bound the total variation between $\hatP$ and $P$.}

Here we bound the total variation between the distributions $\hatP$ and $P$. In fact, we upper bound it in two different ways. Here we present the first way as follows.
\begin{align}
    &\tv(\hatP,P)\nonumber\\
    &\quad =\frac{1}{2}\sum_{x_{1:L+1}\in\calX^{L+1}}\big| \hatP(X_{1:L+1}=x_{1:L+1})
    \\
    &\hphantom{\quad =\frac{1}{2}\sum_{x_{1:L+1}\in\calX^{L+1}}\big|}
    -P(X_{1:L+1}=x_{1:L+1})\big|\nonumber\\
    &\quad = \frac{1}{2}\sum_{x_{1:L}\in\calX^{L}}\big| \hatP(X_{1:L}=x_{1:L})-P(X_{1:L}=x_{1:L})\big|\nonumber\\
    &\quad\leq \frac{1}{2}\sum_{i=0}^{L-1}\sum_{x_{1:i+1}\in\calX^{i+1}}Q(x_{1:i})\prod_{k=1}^{i}f_{k}(x_{1:k},P,Q)\\
    &\quad\quad\cdot \bigg|Q(x_{i+1}\mid  x_{1:i})\cdot f_{i+1}(x_{1:i+1},P,Q)\nonumber\\
    &\quad\quad\quad -P(x_{i+1}\mid  x_{1:i})+G_{i+1}(x_{i+1}\mid x_{1:i},P,Q)\\
    &\quad\qquad\cdot\!\!\sum_{\tilx_{i+1}\in\calX}\!\!\Big[1-f_{i+1}\big((x_{1:i},\tilx_{i+1}),P,Q\big)\Big]Q(\tilx_{i+1}| x_{1:i})\bigg|,\label{ieq:bd1}
\end{align}
where the first equality results from the definition of total variation, the second equality results from Eqn.~\eqref{eq:condp}, the inequality results from Eqn.~\eqref{eq:recur}, and we define 
\begin{align*}
    Q(x_{1:0})\prod_{k=1}^{0}f_{k}(x_{1:k},P,Q)&=1,\quad\\
    Q(\tilx_{1}\mid  x_{1:0})&=Q(\tilx_{1}),\quad \\
    P(\tilx_{1}\mid  x_{1:0})&=P(\tilx_{1})\text{ for all }\tilx_1,x_1\in\calX.
\end{align*}
For each term in the right-hand side of Eqn.~\eqref{ieq:bd1}, we make use of the following proposition.
\begin{proposition}\label{prop:LP}
    For any distributions $P,Q\in\calP(\calX)$ on a finite set $\calX$ and a function $f:\calX\rightarrow\bbR$ such that $1\geq f(x)\geq \min\{1,P(x)/Q(x)\}$ for all $x\in\calX$, we have that
    \begin{align*}
        &\inf_{G\in\calP(\calX)}\!\sum_{x\in\calX}\!\bigg|\!P(x)\!-\!Q(x)f(x)\!-\!G(x)\!\!\sum_{\tilx\in\calX}\!\big(1-f(\tilx)\big)Q(\tilx)\bigg| \\
        &\quad = \sum_{x\in\calX}\big|P(x)-Q(x)\cdot f(x)\big|-\sum_{x\in\calX}\big(1-f(x)\big)Q(x),
    \end{align*}
    and one solution $G^{*}\in\calP(\calX)$ achieving this infimum is
    \begin{align*}
        G^{*} = \nor\big([P-Q\cdot f]_{+}\big).
    \end{align*}
\end{proposition}
The proof of this proposition is provided in Appendix~\ref{app:LP}. By setting the conditional distributions as the distributions in Proposition~\ref{prop:LP}, we can derive one of the optimal solution $G^{*}$ for $f$ as 
\begin{align*}
    &G_{i+1}^{*}(\cdot\mid x_{1:i},P,Q)=\\
    &\quad\nor\Big(\big[ P(\cdot| x_{1:i})-Q(\cdot|  x_{1:i})\cdot f_{i+1}\big((x_{1:i},\cdot),P,Q\big)\big]_{+}\Big).
\end{align*}
The corresponding value of Eqn.~\eqref{ieq:bd1} can be derived as
\begin{align}
    &\tv(\hatP,P)\nonumber\\
    &\quad\leq \frac{1}{2}\sum_{i=0}^{L-1}\sum_{x_{1:i}\in\calX^{i}}Q(x_{1:i})\prod_{k=1}^{i}f_{k}(x_{1:k},P,Q)\\
    &\quad\cdot \!\!\bigg[\sum_{x_{i+1}\in\calX} \!\!\!\!\!\big|P(x_{i+1}| x_{1:i})\!-\!Q(x_{i+1}| x_{1:i})f_{i+1}(x_{1:i+1},P,Q)\big|\nonumber\\
    &\quad -\sum_{x_{i+1}\in\calX}\big(1-f_{i+1}(x_{1:i+1},P,Q)\big)Q(x_{i+1}\mid x_{1:i})\bigg].\label{ieq:bd2}
\end{align}

\textbf{Step 3: Derive the expression of the number of accepted tokens.}

We would like to derive the number of accepted tokens. We define the indicator of whether the $i$-th proposed token is accepted as $I_{i}=\bbI\{\tau\geq i\}$. Then we have that
\begin{align*}
    \tau = \sum_{i=1}^{L}I_{i},\text{ and }\bbE[I_{i}]=\sum_{\tilx_{1:i}\in\calX^{i}}Q(\tilx_{1:i})\prod_{j=1}^{i}f_{j}(\tilx_{1:j},P,Q).
\end{align*}
Thus, the expectation of the number of accepted tokens is
\begin{align*}
    \bbE[\tau+1]=1+\sum_{i=1}^{L}\sum_{\tilx_{1:i}\in\calX^{i}}Q(\tilx_{1:i})\prod_{j=1}^{i}f_{j}(\tilx_{1:j},P,Q).
\end{align*}
We conclude the proof of Theorem~\ref{thm:t_tvbd}.

\section{Proof of Proposition~\ref{prop:opt_resample}}\label{app:resample_proof}
Conditioning on any $x_{1:i}$, we partition the alphabet into two disjoint sets as
\begin{align*}
    \calX^{+}=\{x\,|\, P(x\,|\, x_{1:i})\geq Q(x\,|\, x_{1:i})\}\quad \calX^{-} = \calX-\calX^{+}.
\end{align*}
We then evaluate $G_i^{*}$ and $G_i^{\van}$  on $\calX^{+}$ and $\calX^{-}$. For any $x\in\calX^{-}$, we have that
\begin{align}
     &\big[P(x\,|\,\pt,x_{1:i-1})-Q(x\,|\,\pt,x_{1:i-1})\big]_{+} 
    = 0
    \label{eq:zero}
    \\&
     \big[ P(\cdot\mid x_{1:i})-Q(\cdot\mid  x_{1:i}) f_{i+1}\big((x_{1:i},\cdot),P,Q\big)\big]_{+}= 0,
\end{align}
where the second equality results from the fact that $f_{i}(\cdot \,|\, x_{1:i},P,Q)\geq f_{i}^{\van}(\cdot \,|\, x_{1:i},P,Q)$. For any $x\in\calX^{-}$, we have that
\begin{align*}
    1\geq f_{i}(x\,|\, x_{1:i},P,Q)\geq f_{i}^{\van}(x \,|\, x_{1:i},P,Q)=1,
\end{align*}
where the first inequality results from the fact that $f_{i}$ is the rejection probability. Thus, $f_{i}(x\,|\, x_{1:i},P,Q)$ on $\calX^{-}$. As a result, we have that
\begin{align}
    &\big[ P(\cdot\mid x_{1:i})-Q(\cdot\mid  x_{1:i}) f_{i+1}\big((x_{1:i},\cdot),P,Q\big)\big]_{+}
    \\
    &\quad 
    =\big[P(x\,|\,\pt,x_{1:i-1})-Q(x\,|\,\pt,x_{1:i-1})\big]_{+}.\label{eq:equal}
\end{align}
Combining Eqn.~\eqref{eq:zero} and \eqref{eq:equal}, we have that
\begin{align*}
    &G_{i+1}^{*}(\cdot\mid x_{1:i},P,Q)
    \\&\quad=\nor\Big(\big[ P(\cdot\mid x_{1:i})
    \\
    &\hphantom{\quad=\nor\Big(}
    -Q(\cdot\mid  x_{1:i})\cdot f_{i+1}\big((x_{1:i},\cdot),P,Q\big)\big]_{+}\Big)\nonumber\\
    &\quad=\nor\Big(\big[ P(\cdot\mid x_{1:i})-Q(\cdot\mid  x_{1:i})\big]_{+}\Big)\nonumber\\
    &\quad=G_i^{\van}(\cdot \,|\, x_{1:i-1},P,Q\big).
\end{align*}
Thus, we conclude the proof of Proposition~\ref{prop:opt_resample}.

\section{The Formal Statement and Proof of Proposition~\ref{prop:two_token_informal}}\label{app:formal_proof}
\subsubsection{The Formal Statement of Proposition~\ref{prop:two_token_informal}}
We first state the assumptions for the proposition. To state the assumption, we define the set induced by the acceptance probability functions $f_1,f_2$ as
\begin{align*}
    \calX_1^{+}(f_1)&=\big\{x\,|\, P(x)\geq Q(x) f_1(x,P,Q)\big\},
    \\
    \calX_1^{-}(f_1)&=\calX-\calX_1^{+}(f_1),\nonumber\\
    \calX_2^{+}(x_1,f_2)&=\Big\{x| P(x|x_{1})\geq Q(x|x_{1}) f_2\big((x_{1},x),P,Q\big)\Big\},
    \\
    \calX_2^{-}(x_1,f_2)&=\calX-\calX_2^{+}(x_1,f_2).
\end{align*}
Then we state the following assumption.
\begin{assumption}\label{assump:far}
    The acceptance probability functions $f_1,f_2$ are not close to their unbiased versions $f_1^{\van},f_2^{\van}$. Specifically, we require the following holds for the perturbation constants $c_1$ and $c_2$
    \begin{align*}
        &\bigg|\sum_{x_1\in\calX_{1}^{+}(f_1)}Q(x_1)-\sum_{x_1\in\calX_{1}^{-}(f_1)}Q(x_1)\bigg| = O(c_1+c_2),\\
        &\bigg|\sum_{x_2\in\calX_2^{-}(x_1,f_2)}Q(x_2\,|\,x_1)-\!\!\!\!\!\!\sum_{x_2\in\calX_2^{+}(x_1,f_2)}\!\!Q(x_2\,|\,x_1)\bigg|  
        \\*
        &\quad  = O(c_1+c_2)\quad\text{ for all }x_1.
    \end{align*}
\end{assumption}
For $f_1^{\van}$ and $f_2^{\van}$, the left-hand sides of these two equalities are all equal to $1$. In contrast, we require their values to be upperbounded by $O(c_1+c_2)$ for $f_1,f_2$, which is strictly smaller than $1$.
\begin{assumption}\label{assump:closeness}
    The target model $P$ and the draft model $Q$ are close, i.e.,
    \begin{align*}
        &\tv\big(P(\cdot\,|\, \texttt{prefix}),Q(\cdot\,|\, \texttt{prefix})\big)
        \\
        &\quad=\frac{1}{2}\sum_{x\in\calX}\big|P(x\,|\, \texttt{prefix})-Q(x\,|\, \texttt{prefix})\big|\leq \frac{2}{5}
    \end{align*}
    for all $\texttt{prefix}\in\calX^{*}$.
\end{assumption}
Here the number $2/5$ can be any value that is smaller than $1/2$. The draft models we trained with target model being LlamaGen-XL~\citep{sun2024autoregressive} and Lumina-mGPT~\citep{liu2024lumina} can achieve TV in Table~\ref{tab:tv}, where we obtained the result on $1000$ images generated according to randomly selected prompts from MS-COCO 2017 training set~\citep{lin2014microsoft}. This empirically verifies our assumption. 

\begin{table}[]
\centering
\begin{tabular}{|c|c|c|}
\hline 
Model       & LlamaGen-XL & Lumina-mGPT \\ \hline
TV distance & $0.38$        & $0.32$     \\ \hline
\end{tabular}
\caption{The TV distance between the target and draft models of LlamaGen-XL and Lumina-mGPT.} \label{tab:tv}
\end{table}

In the following, we conduct the perturbation analysis of any general $\{f_i\}_{i=1}^{L}$ when $L=2$ as
\begin{align*}
    \tilf_{1}(x_1, c_1, P,Q)&=f_1(x_1,P,Q)+c_1, \\
    \tilf_{2}((x_{1},x_{2}),c_2, P,Q)&=f_2((x_{1},x_{2}),P,Q)+c_2.
\end{align*}
Here we require that $|c_1|,|c_2|=o(1)$, i.e., the amplitudes of the perturbations are small as the common perturbation analysis~\citep{bonnans2013perturbation}. We denote the total variation bound, i.e., the right-hand side of Eqn.~\eqref{ieq:tv_token} $\{\tilf_i\}_{i=1}^{2}$ as $\texttt{TVB}(c_1,c_2)$.
\begin{proposition}\label{prop:two_token}
    Maintaining the same expectation of the number of accepted tokens for $\{\tilf_i\}_{i=1}^{2}$ and $\{f_i\}_{i=1}^{2}$ requires that
    \begin{align}
        c_1 \!= \!\frac{-\bbE_{X_1\sim Q}\big[f_1(X_1,P,Q)\big]}{1\!+\!\bbE_{(X_1,X_2)\sim Q}\Big[f_2\big((X_1,X_2),P,Q\big)\Big]\!+\!c_2}\!\cdot c_2\label{eq:c1c2_stat}.
    \end{align}
    If Assumptions~\ref{assump:far} and \ref{assump:closeness} hold and $f_2^{\van}((x_1,x_2),P,Q)\leq f_2((x_1,x_2),P,Q)$ pointwisely, we have the following for two perturbations $(c_1,c_2)$ and $(\tilc_1,\tilc_2)$ satisfying Eqn.~\eqref{eq:c1c2_stat} but with different signs, i.e., $c_1,\tilc_2 >0$, $c_2,\tilc_1 <0$,
    \begin{align*}
        \texttt{TVB}(c_1,c_2)<\texttt{TVB}(\tilc_1,\tilc_2).
    \end{align*}
\end{proposition}
In the following, we prove this proposition.
\subsubsection{Proof Proposition~\ref{prop:two_token}}
Here we conduct the perturbation analysis of $f_1$ and $f_2$ as $\tilf_{1}(x_1,P,Q)=f_1(x_1,P,Q)-c_1$ and $\tilf_{2}((x_{1},x_{2}),P,Q)=f_2((x_{1},x_{2}),P,Q)+c_2$ for all $x_1,x_2\in\calX$, respectively. The constants $c_1$ and $c_2$ can be positive or negative. We adopt different signs for $f_1$ and $f_2$ just for mathematical convenience. According to Eqn.~\eqref{eq:token_exp}, the expectation of the number of accepted tokens for $\tilf_1$ and $\tilf_2$ is
\begin{align*}
    &\sum_{x_1\in\calX}Q(x_1)\tilf_1(x_1,P,Q)
    \\*
    &\quad
    +\sum_{(x_1,x_2)\in\calX^2}Q(x_1,x_2)\tilf_1(x_1,P,Q)\tilf_2((x_{1},x_{2}),P,Q)\nonumber\\
    &\; = \bbE_{X_1\sim Q}\big[f_1(X_1,P,Q)\big]
    \\
    &\quad\quad
    +\bbE_{(X_1,X_2)\sim Q}\Big[f_1(X_1,P,Q)\cdot f_2\big((X_1,X_2),P,Q\big)\Big]\nonumber\\
    &\quad\quad+c_2\cdot \bbE_{X_1\sim Q}\big[f_1(X_1,P,Q)\big]
    \\
    &
    \quad\quad
    -\bigg(1+\bbE_{(X_1,X_2)\sim Q}\Big[f_2\big((X_1,X_2),P,Q\big)\Big]\bigg)c_1
    \\
    &
    \quad\quad
    -c_1\cdot c_2.
\end{align*}
To keep the expectations of the numbers of accepted tokens the same for $\{f_i\}_{i=1}^{2}$ and $\{\tilf_i\}_{i=1}^{2}$, we require that
\begin{align}
    c_1 = \frac{\bbE_{X_1\sim Q}\big[f_1(X_1,P,Q)\big]}{1\!+\!\bbE_{(X_1,X_2)\sim Q}\Big[f_2\big((X_1,X_2),P,Q\big)\Big]\!+\!c_2}\!\cdot\! c_2.\label{eq:c1c2}
\end{align}
We consider the regime where $c_1,c_2\ll 1$. Thus, Eqn.~\eqref{eq:c1c2} implies that they should have the same sign. Since we set the resampling distribution as $G^{*}$ in Eqn.~\eqref{eq:opt_G}, the right-hand side of Eqn.~\eqref{ieq:tv_token} becomes
\begin{align}
    &b(f_1,f_2) 
    \\
    &\quad = \sum_{x_1\in\calX}\big|P(x_1)-Q(x_1)f_1(x_1,P,Q)\big|
    \\
    &\quad\quad-\sum_{x_1\in\calX}Q(x_1)\big[1-f_1(x_1,P,Q)\big] \nonumber\\
    &\quad\quad +\sum_{x_1\in\calX}Q(x_1)f_1(x_1,P,Q)
    \\
    &\quad\quad\;
    \cdot\bigg[\sum_{x_2\in\calX}\Big|P(x_{2}\,|\,x_{1})-Q(x_{2}\,|\,x_{1})f_2\big((x_1,x_2),P,Q\big)\Big|\nonumber\\
    &\quad \qquad\quad -\sum_{x_2\in\calX}Q(x_{2}\,|\,x_{1})\Big[1-f_2\big((x_1,x_2),P,Q\big)\Big]\bigg]
\end{align}
To facilitate the analysis of $b(\tilf_1,\tilf_2)$, we define four sets as follows.
\begin{align*}
    \calX_1^{+}(f_1)&=\big\{x\,|\, P(x)\geq Q(x) f_1(x,P,Q)\big\},
    \\
    \calX_1^{-}(f_1)&=\calX-\calX_1^{+}(f_1),\nonumber\\
    \calX_2^{+}(x_1,f_2)&=\Big\{x\,|\, P(x\,|\,x_{1})\geq Q(x\,|\,x_{1}) f_2\big((x_{1},x),P,Q\big)\Big\},
    \\
    \calX_2^{-}(x_1,f_2)&=\calX-\calX_2^{+}(x_1,f_2).
\end{align*}
Some basic calculations lead to
\begin{align}
    &b(\tilf_1,\tilf_2)\nonumber\\
    &\quad=b(f_1,f_2)+\bigg[\Big(\sum_{x_1\in\calX_{1}^{+}(f_1)}Q(x_1)-\sum_{x_1\in\calX_{1}^{-}(f_1)}Q(x_1)\Big)
        \\
        &\hphantom{\quad=b(f_1,f_2)+\bigg[}
        -1-\Delta(f_2,P,Q)\bigg] c_1\nonumber\\
        &\quad +\sum_{x_1\in\calX}Q(x_1)f_{1}(x_1,P,Q)\bigg[\bigg(\sum_{x_2\in\calX_2^{-}(x_1,f_2)}Q(x_2\,|\,x_1)
        \\
        &\quad\qquad 
        -\!\!\!\!\sum_{x_2\in\calX_2^{+}(x_1,f_2)}\!\!Q(x_2\,|\,x_1)\bigg)+1\bigg]c_2 +O(c_1\cdot c_2)\nonumber\\
    &\quad = b(f_1,f_2) + \sum_{x_1\in\calX}Q(x_1)f_{1}(x_1,P,Q) c_2 
    \\
    &\quad\qquad
    - (1+\Delta(f_2,P,Q))c_1 +O(c_1^2+c_2^2+c_1c_2),\label{eq:delta_b_1}
\end{align}
where we define the quantity
\begin{align*}
    &\Delta(f_2,P,Q) \\
    &= \sum_{x_1\in\calX}\!\!Q(x_1)\Big[\sum_{x_{2}\in\calX}\!\big|P(x_2\,|\,x_1)
    \\
    &\quad\qquad\qquad
    -Q(x_2\,|\,x_1)f_2((x_1,x_2),P,Q)\big|\!
     \\
    &\quad\qquad\qquad
    -\!\!\!\!\sum_{x_2\in\calX}\!Q(x_2\,|\,x_{1})\big[1-f_2((x_1,x_2),P,Q)\big]\Big],
\end{align*}
and the second equality results from Assumption~\ref{assump:far}. Plugging Eqn.~\eqref{eq:c1c2} into Eqn.~\eqref{eq:delta_b_1}, we have that
\begin{align}
    &b(\tilf_1,\tilf_2)\nonumber\\
    &\quad = b(f_1,f_2) + \bbE_{X_1\sim Q}\big[f_1(X_1,P,Q)\big]\cdot 
    \\
    &\quad
    \frac{\bigg(\bbE_{(X_1,X_2)\sim Q}\Big[f_2\big((X_1,X_2),P,Q\big)\Big]\!-\!\!\Delta(f_2,\!P,Q)\!-\!c_2\!\bigg)\!c_2}{1+\bbE_{(X_1,X_2)\sim Q}\Big[f_2\big((X_1,X_2),P,Q\big)\Big]+c_2} \nonumber\\
    &\quad\qquad +O(c_1^2+c_2^2+c_1c_2).\label{eq:plug}
\end{align}
To decide whether $c_2$ should be positive or negative, we have the following proposition.
\begin{proposition}\label{prop:positive}
    If Assumption~\ref{assump:closeness} holds and $f_2^{\van}\leq f_{2}$ pointwisely, we have that
    \begin{align*}
        \bbE_{(X_1,X_2)\sim Q}\Big[f_2\big((X_1,X_2),P,Q\big)\Big]\geq \Delta(f_2,P,Q)+\frac{1}{5}.
    \end{align*}
\end{proposition}
The proof of this proposition is in Appendix~\ref{app:positive_proof}. Thus, to minimize the total variation upper bound, $c_2$ is negative. We conclude the proof of Proposition~\ref{prop:two_token}.

\section{Proof of Proposition~\ref{prop:positive}}\label{app:positive_proof}
In order to prove the desired result, we prove a sufficient condition for it, which is
\begin{align*}
    &\frac{1}{5}+\sum_{x_{2}\in\calX}\!\big|P(x_2\,|\,x_1)-Q(x_2\,|\,x_1)f_2((x_1,x_2),P,Q)\big|\!
    \\
    &\quad
    -\!\!\sum_{x_2\in\calX}\!Q(x_2\,|\,x_{1})\big[1-f_2((x_1,x_2),P,Q)\big]\\
    &\quad \leq \sum_{x_{2}\in\calX} Q(x_2\,|\,x_1)f_2\big((x_1,x_2),P,Q\big)\text{ for all }x_1\in\calX.
\end{align*}
We note that this is equivalent to
\begin{align*}
    \sum_{x_{2}\in\calX}\!\big|P(x_2\,|\,x_1)-Q(x_2\,|\,x_1)f_2((x_1,x_2),P,Q)\big|\leq \frac{4}{5}.
\end{align*}
For each term in this summand, we have that
\begin{align*}
    & \big|P(x_2\,|\,x_1)-Q(x_2\,|\,x_1)f_2((x_1,x_2),P,Q)\big|\\
    &\quad = \big|P(x_2\,|\,x_1)-Q(x_2\,|\,x_1)f_2((x_1,x_2),P,Q)\big|
        \\
        &\quad\qquad\quad
        \cdot \bbI\{P(x_2\,|\,x_1)\geq Q(x_2\,|\,x_1)\}\\
        &\quad\qquad +\big|P(x_2\,|\,x_1)-Q(x_2\,|\,x_1)f_2((x_1,x_2),P,Q)\big|
        \\
        &\quad\qquad\quad
        \cdot \bbI\{P(x_2\,|\,x_1)< Q(x_2\,|\,x_1)\}\\
    &\quad \leq \big(P(x_2|x_1)-Q(x_2|x_1)\big)\cdot \bbI\{P(x_2\,|\,x_1)\geq Q(x_2\,|\,x_1)\}\\
    &\quad\qquad +\big(Q(x_2\,|\,x_1) - P(x_2\,|\,x_1)\big)
    \\
        &\quad\qquad\quad
        \cdot \bbI\{P(x_2\,|\,x_1)< Q(x_2\,|\,x_1)\}\\
    &\quad = |P(x_2\,|\,x_1)-Q(x_2\,|\,x_1)\big|,
\end{align*}
where the inequality results from the fact that $f_2^{\van}\leq f_{2}\leq 1$ pointwisely. Thus, we only need to prove
\begin{align*}
    \sum_{x_{2}\in\calX}\!|P(x_2\,|\,x_1)-Q(x_2\,|\,x_1)\big|\leq \frac{4}{5}.
\end{align*}
This is exactly Assumption~\ref{assump:closeness}. Thus, we conclude the proof of Proposition~\ref{prop:positive}.
\section{Proof of Supporting Propositions}
\subsection{Proof of Proposition~\ref{prop:LP}}\label{app:LP}
The proof consists of two parts. In the first part, we show that the lower bound in Proposition~\ref{prop:LP} is a valid lower bound. In fact, for any $G\in\calP(\calX)$, we have that
\begin{align*}
    &\sum_{x\in\calX}\bigg|P(x)-Q(x)\cdot f(x)-G(x)\sum_{\tilx\in\calX}\big(1-f(\tilx)\big)Q(\tilx)\bigg|\\
    &\quad\geq \sum_{x\in\calX}\big|P(x)-Q(x)\cdot f(x)\big|
    \\
        &\quad\qquad-\sum_{x\in\calX}G(x)\bigg|\sum_{\tilx\in\calX}\big(1-f(\tilx)\big)Q(\tilx)\bigg|\\
    &\quad = \sum_{x\in\calX}\big|P(x)-Q(x)\cdot f(x)\big| - \sum_{\tilx\in\calX}\big(1-f(\tilx)\big)Q(\tilx),
\end{align*}
where the inequality results from the triangle inequality, and the equality results from the facts that $G$ is a distribution and $0\leq f(x)\leq 1$ for all $x\in\calX$. Then we show that the distribution $G^{*}$ in Proposition~\ref{prop:LP} can achieve this. We only consider the case $\sum_{\tilx\in\calX}\big(1-f(\tilx)\big)Q(\tilx)>0$. Otherwise, any $G$ can achieve the infinimum. In fact, we have that
\begin{align}
    0&< \sum_{\tilx\in\calX}\big(1-f(\tilx)\big)Q(\tilx) 
    \\
    &= \sum_{\tilx\in\calX}P(\tilx)-f(\tilx)Q(\tilx)
    \\
    &\leq \sum_{\tilx\in\calX}\big[P(\tilx)-f(\tilx)Q(\tilx)\big]_{+},\label{ieq:sum}
\end{align}
where the equality results from the normalization of distributions. 
In the second part, we define two disjoint subsets of $\calX$ as 
\begin{align*}
    &\calX_{+}=\{x\in\calX\,|\, P(x)> Q(x)\cdot f(x)\},
    \\*
    &\calX_{-}=\{x\in\calX\,|\, P(x)\leq Q(x)\cdot 
    f(x)\}.
\end{align*}
Then $\calX=\calX_{+}\cup\calX_{-}$. Thus, we have that
\begin{align*}
    &\sum_{x\in\calX}\bigg|P(x)-Q(x)\cdot f(x)-G^{*}(x)\sum_{\tilx\in\calX}\big(1-f(\tilx)\big)Q(\tilx)\bigg|\\
    &\quad = \sum_{x\in\calX}\bigg|P(x)-Q(x)\cdot f(x)
        \\
        &\quad\qquad
        -\!\frac{\big[P(x)\!-\!f(x)Q(x)\big]_{+}}{\sum_{\tilx\in\calX}\!\big[P(\tilx)\!-\!f(\tilx)Q(\tilx)\big]_{+}}\sum_{\tilx\in\calX}\big(1\!-\!f(\tilx)\big)Q(\tilx)\bigg|\\
    &\quad =\sum_{x\in\calX_{+}}P(x)-Q(x)\cdot f(x)
        \\
        &\quad\qquad
        -\frac{\big[P(x)\!-\!f(x)Q(x)\big]_{+}}{\sum_{\tilx\in\calX}\big[P(\tilx)\!-\!f(\tilx)Q(\tilx)\big]_{+}}\sum_{\tilx\in\calX}\big(1-f(\tilx)\big)Q(\tilx)\\
        &\quad\qquad + \sum_{x\in\calX_{-}}\big|P(x)-Q(x)\cdot f(x)\big|\\
    &\quad = \sum_{x\in\calX}\big|P(x)-Q(x)\cdot f(x)\big|-\sum_{\tilx\in\calX}\big(1-f(\tilx)\big)Q(\tilx),
\end{align*}
where the second equality results from Eqn.~\eqref{ieq:sum}. Thus, we conclude the proof of Proposition~\ref{prop:LP}.


\end{document}